\documentclass[letterpaper,journal]{IEEEtran}
\usepackage{amsmath,amsfonts}
\usepackage{algorithmic}
\usepackage{algorithm}
\usepackage{array}
\usepackage[caption=false,font=normalsize,labelfont=sf,textfont=sf]{subfig}
\usepackage{textcomp}
\usepackage{stfloats}
\usepackage{url}
\usepackage{verbatim}
\usepackage{graphicx}
\usepackage{cite}
\usepackage{booktabs} 
\usepackage{multirow} 
\usepackage{balance}

\begin{document}

\title{Graph-Structured Hyperdimensional Computing for Data-Efficient and Explainable Process–Structure–Property Prediction}

\author{Jingzhan~Ge, Ajeeth~Vellore, Ajinkya~Palwe, Ahsan~Khan, David~Gorsich, Matthew~P.~Castanier, SeungYeon~Kang, and Farhad~Imani%
\thanks{Manuscript received Month Day, Year; revised Month Day, Year. (Corresponding author: Farhad Imani.)}%
\thanks{J. Ge, A. Vellore, A. Palwe, A. Khan, S. Kang, and F. Imani are with the School of Mechanical, Aerospace, and Manufacturing Engineering, University of Connecticut, Storrs, CT 06269 USA (e-mail: farhad.imani@uconn.edu).}%
\thanks{D. Gorsich and M. P. Castanier are with the U.S. Army Combat Capabilities Development Command, Ground Vehicle Systems Center, Warren, MI 48397 USA.}%
\thanks{DISTRIBUTION STATEMENT A. Approved for public release; distribution is unlimited. OPSEC10422}%
\markboth{Journal of \LaTeX\ Class Files,~Vol.~14, No.~8, August~2021}%
{Shell \MakeLowercase{\textit{et al.}}: A Sample Article Using IEEEtran.cls for IEEE Journals}
}

\maketitle

\begin{abstract}
Multiphoton photoreduction enables high-fidelity fabrication of complex 3D microstructures, yet reliable process–structure–property (PSP) prediction remains difficult because the available data are sparse, heterogeneous, and interaction-dominated. In this regime, conventional feature-vector models are statistically underdetermined, making them prone to spurious correlations, poor regime transfer, and unstable post hoc explanations, whereas mechanistic pipelines depend on calibrated submodels that are rarely available during early process development. We present PSP-HDC, a graph-structured hyperdimensional computing framework that encodes a directed PSP graph as an internal prior for representation, inference, and explanation. A trainable scalar-to-hypervector encoder learns parameter-specific embeddings on a fixed hyperdimensional basis to accommodate heterogeneous scales and noise. Sample representations are then composed through graph-aligned binding and bundling along directed PSP dependencies, and prediction is performed by associative-memory retrieval against class prototypes. Because the same prototype memories support both decision making and attribution, PSP-HDC provides intrinsic explanations at the parameter, group, and within-group levels, while memory alignment and separation quantify prototype formation during training. On sheet-resistance regime prediction for the 3D platform, PSP-HDC achieves an accuracy of $0.910 \pm 0.077$ over 1000 random splits and $0.896$ under process-fold generalization, outperforming strong baselines.

\end{abstract}

\def\abstractname{Note to Practitioners}
\begin{abstract}
Building reliable process–structure–property (PSP) models for advanced manufacturing is difficult because each labeled specimen is expensive, requiring a controlled build along with microscopy, post-processing, and property testing. With only dozens of heterogeneous measurements, black-box predictors often overfit and their explanations are unstable. PSP-HDC is a lightweight decision-support model for this data-scarce setting. Practitioners specify a directed PSP graph (process to structure to property) and simple variable groups; the model learns calibrated signatures for each scalar measurement and combines them only through the graph’s allowed dependencies. A new build is classified by similarity to learned class prototypes, and the same prototypes provide built-in attributions that highlight which process knobs and structure descriptors are driving the decision, useful for process-window screening and experiment prioritization. PSP-HDC retrains quickly as new experiments arrive; the similarity margin can be monitored as a practical confidence indicator. The method is not a substitute for mechanistic validation. It relies on a well-constructed PSP graph and physically meaningful structure descriptors, and it must be re-tuned whenever the sensing or measurement pipeline changes. Beyond AM, the approach can support other automation problems where data are scarce but dependency structure is known.
\end{abstract}


\section{Introduction}
\IEEEPARstart{P}{rocess}--structure--property (PSP) prediction becomes especially difficult in data-scarce laser-based fabrication because small changes in processing conditions can alter local reduction behavior, microstructure, and final electrical performance through strongly coupled pathways. In the 3D OHMIC setting studied here, the resulting PSP mapping is high dimensional and interaction dominated. In parallel, fully characterized samples are expensive; each data point requires controlled fabrication, microscopy, segmentation, chemical-state characterization, and electrical testing, producing small and heterogeneous PSP tables \cite{Fang2024TASE_AMMonitoringSurvey, Tran2023TASE_IAMA_LPBF}. In this regime, increasing model capacity alone does not ensure reliable prediction or stable interpretation \cite{mu2024online}. Robust learning instead demands explicit structural constraints that encode admissible interactions and preserve the directed dependence from process to structure to property \cite{liu2024interpretable}.

Without an explicit structural prior, the two dominant PSP modeling routes both break under the combination of strong coupling and small data because interaction topology is underdetermined by the observations. The first route is modular physics or multiphysics chaining, where process inputs drive thermal or kinetic models that then feed structure evolution and property models \cite{yan2018integrated}. In multiphoton photoreduction, photoreduction kinetics, optical interactions, transport, interfaces, and defect formation are tightly coupled; submodels tuned in isolation are therefore difficult to transfer and their errors compound when chained \cite{Guo2023TASE_LMD_ThermalSignature}. The second route is data-driven prediction that treats PSP inputs as a feature vector and fits a regressor or classifier directly from experimental tuples \cite{luo2024dataset}. This route fails for a statistical rather than capacity reason because sparse coverage allows many incompatible high-order interaction patterns to fit the same samples, causing regime-specific correlations, poor extrapolation, and unstable post-hoc explanations \cite{saunders2023metal, Wang2022TASE_ShapeDeviation_AM, ye2023surrogate}.

Graph-based artificial intelligence targets this failure mode by encoding relational structure as an internal representation rather than as an optional annotation. In mainstream message-passing/attention graph neural networks (e.g., GCN and GraphSAGE), the neighbor-aggregation/edge functions are learned from data \cite{kipf2017semi,Alenezi2025TASE_PWL}. Attention-based variants also learn edge-specific weights \cite{velickovic2018graph}. In this setting, each specimen is a small, heterogeneous PSP table and the relevant directed PSP graph is shared across samples; with tens of samples, learning edge weights, attention patterns, or latent adjacency is statistically underdetermined and produces brittle, regime-specific couplings. PSP has a directed topology in which process influences structure, and property depends on the joint configuration of process and structure. Using a domain-defined directed PSP graph as a fixed internal prior constrains admissible compositions, enforces PSP ordering, and aligns explanations with the same dependency paths that generate predictions, producing dependency-aware evidence rather than post-hoc narratives. In the small-data regime, this shift from flat coordinates to constrained relational composition is the difference between interpolation that overfits correlations and inference that generalizes under structural constraints.

To address these constraints, we propose PSP-HDC, a graph-based hyperdimensional computing (HDC) framework for per-sample PSP prediction in small, heterogeneous experimental datasets. HDC, also known as vector-symbolic learning, represents variables as high-dimensional hypervectors and computes with algebraic operations that support robust compositional representations. PSP-HDC instantiates a directed PSP graph from domain knowledge and uses it as the internal scaffold for representation, inference, and explanation. The graph is not a visualization layer. It is the computational prior that dictates how process and structure evidence can be combined and how property decisions are formed. 

Graph constraints are enforced by construction through graph-aligned hypervector composition. Node-level variables are encoded as hypervectors, parameter groups are aggregated by bundling, and directed dependencies are encoded by binding along graph edges so that interaction terms exist only where the graph permits them. This composition converts a heterogeneous PSP table into a constrained relational representation in which process-to-structure and structure-to-property dependencies are explicit. Property prediction is performed by associative-memory retrieval against class prototypes learned from training samples. Prototype retrieval yields a controlled decision rule based on similarity margins.

A key barrier is that the PSP graph nodes are populated by messy experimental scalars with different scales and noise characteristics. Classical HDC encoders based on fixed random projections treat all variables as statistically equivalent. In heterogeneous experimental PSP tables, that assumption is false and destructive. If node representations are poorly calibrated, graph composition propagates distortions across edges and degrades prototype separability, weakening both accuracy and attribution stability. PSP-HDC therefore introduces a trainable scalar-to-hypervector encoder that learns compact parameter-specific embeddings on a fixed high-dimensional basis. It adapts each node representation to the statistics of its measurement pipeline while retaining the efficiency and noise tolerance of HDC. As a result, the graph can impose meaningful constraints and the associative memories can form sharp, separable prototypes even under small data.

PSP-HDC also provides intrinsic explainability that is aligned with the PSP graph rather than reconstructed after prediction. Because class prototypes are built through graph-consistent composition, decision evidence can be decomposed into parameter-level contributions, parameter-group contributions, and within-group contributions. This multi-level attribution is essential for PSP learning where coupled effects dominate and where single-variable rankings are scientifically misleading. Group-level evidence isolates coordinated effects across correlated descriptors, and within-group decompositions resolve which variables dominate inside a coupled set. We further introduce memory alignment and separation (MAS) to quantify learning dynamics at the memory level. MAS measures how class-partitioned component memories at the parameter and group levels align with their corresponding class prototype and separate from competing prototypes during training. Since class prototypes are formed by composing component memories along the directed PSP graph, increases in alignment and separation correspond to sharper prototypes and larger retrieval margins. 

We evaluate PSP-HDC on a small, heterogeneous PSP dataset generated by a femtosecond-laser-based multiphoton photoreduction process~\cite{awasthi2025revolutionizing}. The experiments were conducted on the 3D OHMIC research platform; however, the present study is restricted to specimens fabricated using only its multiphoton photoreduction mode. Using both random splits and process-fold regime holdouts that test exclusively on unseen process regimes, we directly probe whether the directed PSP graph prior captures transferable PSP coupling rather than regime-specific correlations in a small, heterogeneous PSP table. Our contributions are as follows.

\begin{itemize}
\item We introduce a directed graph as an internal structural prior for small-data learning, turning process-to-structure-to-property causality into a computational constraint rather than an external annotation.
\item We develop PSP-HDC, a graph-structured HDC model that enforces PSP topology through graph-aligned binding and bundling and performs prediction by associative-memory retrieval against class prototypes.
\item We design a trainable scalar-to-hypervector encoder with parameter-specific embeddings on a fixed high-dimensional basis, enabling the PSP graph to interface with heterogeneous, noisy experimental scalars while preserving HDC efficiency.
\item We derive intrinsic, graph-aligned explanations with attributions at the parameter, group, and within-group levels from the same prototype memories used for prediction, yielding coupled-effect evidence suitable for planning.
\item We propose memory alignment and separation to quantify prototype formation dynamics across the PSP graph and to connect memory geometry to retrieval margins and predictive reliability.

\end{itemize}

The remainder of this paper is organized as follows. Section~\ref{sec_related_work} reviews related work. Section~\ref{sec_methodology} presents the methodology. Section~\ref{sec: Experiment design} describes the experimental design. Section~\ref{sec: Results} reports results and analysis. Section~\ref{sec: Conclusion} concludes and discusses future directions.

\section{Related Work}\label{sec_related_work}

\subsection{Per-Sample PSP Datasets and Characterization Limits}\label{sec: PSP dataset}
Progress in PSP learning is gated by the availability of aligned per-specimen process, structure, and property records, because structure extraction typically requires multimodal metrology and nontrivial post-processing. Two widely used full-PSP benchmarks in AM are the laser powder bed fusion (L-PBF) metal datasets released by Luo \emph{et al.} \cite{luo2023dataset, luo2024dataset}. Their Ti-6Al-4V dataset links process settings to quantified structure descriptors from XCT/SEM/EBSD and reports multiple mechanical properties across 42 distinct process conditions, with one specimen per condition. Their AlSi10Mg dataset provides a similar per-sample PSP record across 60 conditions, but complete grain/sub-grain descriptors are available only for a subset of conditions, reducing the number of specimens with fully populated structure descriptors.

Most other public resources either provide only partial PSP legs or distribute measurements across tasks rather than providing a single consolidated PSP table per specimen. Literature-mined compilations often emphasize process--property relations (e.g., density or strength trends) while omitting structure descriptors needed to resolve mediation and coupling \cite{barrionuevo2025laser}. Structure repositories provide large annotated micrograph collections, but typically lack paired process history and downstream properties, limiting their value for process optimization \cite{iren2021aachen}. AM-Bench releases rich measurements, yet they are frequently organized as benchmark tasks or distributed measurements rather than a unified per-specimen PSP record \cite{heigel2020ambench}. Related datasets outside AM can match the PSP paradigm (process--microstructure--mechanical response), but they do not address AM PSP learning and associated regime-shift questions \cite{salvador2022ti_dataset}.

These dataset characteristics imply two practical consequences for PSP learning. First, complete PSP tables often sit in a small-$N$, heterogeneous-statistics regime because the cost of structure acquisition and validation dominates throughput. Second, the measured structure leg is rarely a single modality; it is typically a composite of descriptors produced by different imaging pipelines and segmentation heuristics, which creates parameter-dependent noise and scale differences even before any learning begins. The 3D OHMIC setting studied here remains highly heterogeneous because specimens fabricated via multiphoton photoreduction are paired with multimodal microscopy- and SEM-EDS-derived structure descriptors and electrical property targets that span orders of magnitude~\cite{awasthi2025revolutionizing}.

\subsection{PSP Prediction in Additive Manufacturing}\label{sec: psp_prediction}
Prior PSP prediction strategies in AM can be organized by what they assume about mechanistic portability and what they assume about statistical identifiability in small full-PSP tables. Physics chains and physics-informed models can be strong when submodels are calibrated and transferable, while purely data-driven models can be strong when data volume and regime coverage are sufficient. The most relevant evidence for the present study comes from settings that expose where these assumptions fail under strong PSP coupling and limited data \cite{fang2024tase_am_survey}.

Multiscale PSP pipelines typically propagate process variables through chained thermal/kinetic models into structure evolution and then property models. This modularity enables interpretability and reuse when each stage remains valid across the operating window, but it demands extensive calibration and can incur substantial computational cost. Surrogate-based variants aim to reduce runtime by generating physics-based data across stages and learning reduced-order mappings \cite{guo2023tase_lmd_surrogate}. For example, Saunders \emph{et al.} build a hybrid surrogate framework for L-PBF 316L by generating physics data across multiple stages and training Gaussian Process surrogates \cite{saunders2023metal, ye2023surrogate}. Such approaches can be effective when the chained physics is well validated, but they still inherit the portability limitations of their prerequisite models and can accumulate stage-wise error when coupling or boundary conditions change.

Physics-informed learning injects physical knowledge through features, architectural constraints, or training objectives \cite{alenezi2025pwl}. Evidence suggests these methods are most effective when (i) the governing structure is well specified, or (ii) dense sensor streams strongly constrain intermediate states that dominate the target property. For instance, Ghungrad \emph{et al.} constrain a deep model by deep unfolding of transient heat-transfer structure for thermal-history prediction \cite{ghungrad2023architecture}. Cooper \emph{et al.} use physics-motivated preprocessing on in-situ thermal histories for tensile prediction, where engineered thermal derivatives act as proxies for solidification-relevant cues \cite{cooper2023tensile}. McGowan \emph{et al.} incorporate simulation-informed loss terms for porosity quality classification from in-situ images, which depends on alignment between simulation cues and observed melt-pool patterns \cite{mcgowan2022physics}. In per-specimen PSP learning with heterogeneous scalar descriptors and incomplete mechanistic calibration, physics-guided proxies can introduce model-form bias and do not necessarily resolve the identifiability of coupled PSP interactions \cite{kats2022physics}.

A large body of work fits regressors/classifiers directly on experimental PSP tuples using classical ML (e.g., RF/SVM/GP) and deep backbones when sufficient training diversity exists. Under small full PSP tables, the key limitation is not capacity but the underdetermination of interaction structure, which creates sensitivity to splitting protocol and correlation patterns \cite{ye2023surrogate}. Luo \emph{et al.} propose bootstrapped augmentation around measured statistics to mitigate data scarcity for PSP prediction \cite{luo2023new}. This strategy increases sample count but does not guarantee physically consistent PSP tuples, and ranking-based explanations can remain unstable under correlated inputs. Liu \emph{et al.} use ARD Gaussian Process surrogates and predictive uncertainty to guide exploration in a small-sample L-PBF setting \cite{liu2024interpretable}. While ARD provides an interpretable inductive bias, length-scale rankings become ambiguous under correlated inputs and fail to capture coordinated effects from coupled PSP mechanisms.

Deep models often succeed by learning surrogates for specific modalities, but the literature shows that these successes can bypass PSP mediation. Fang \emph{et al.} predict tensile properties from thermal histories using CNNs, which is effective when signals concentrate property-relevant information \cite{fang2022data}. Whitman \emph{et al.} embed microstructure images using pretrained vision transformers and regress properties from those embeddings, which yields an efficient structure--property surrogate when microstructure images dominate the mapping \cite{whitman2025machine}. These approaches provide valuable modality-specific surrogates, yet they do not, by construction, resolve how process and structure jointly compose under directed PSP dependency, and their explanations are typically post hoc or embedding-mediated rather than dependency-path aligned.

\subsection{Graph-Based Learning for PSP}\label{sec: graph_based}
Graph-based models encode relational inductive bias by parameterizing interaction through edges and neighborhood aggregation \cite{alenezi2025gvpnet}. In mainstream message-passing GNNs (e.g., GCN/GraphSAGE), neighbor aggregation and edge functions are learned from data \cite{hamilton2017inductive}. Attention-based variants learn edge-specific weights that act as data-driven interaction strengths. Empirically, these models excel when training exposes many graphs and diverse local motifs (e.g., molecules, meshes, sensor networks), allowing edge functions and attention patterns to be identified from repeated structural variation.

Per-sample PSP learning differs in two ways that sharpen the statistical burden. First, each specimen corresponds to a small table of heterogeneous scalar descriptors, not a large homogeneous graph with abundant local neighborhoods. Second, the directed PSP structure is typically shared across specimens, so a learned message-passing model must identify interaction functions from a limited number of repeated realizations of the same topology, which can be underdetermined in the tens-of-samples regime. As a consequence, graph neural architectures alone do not guarantee that learned interactions correspond to stable PSP dependency paths under regime holdouts, and additional constraints are often needed to enforce directed process$\rightarrow$structure$\rightarrow$property ordering rather than allowing shortcut couplings through flexible attention. These observations motivate enforcing graph priors by construction when data are insufficient to reliably learn interaction topology.

\subsection{HDC and Graph-Aware Composition}\label{sec: HDC}
HDC, also known as vector-symbolic architectures, represents entities as high-dimensional hypervectors and performs computation using a small set of compositional operators (binding and bundling) together with similarity-based associative-memory retrieval \cite{kanerva2009hyperdimensional}. The high-dimensional regime supports quasi-orthogonal representations, which yields noise-tolerant similarity search and prototype-based inference with explicit margins that can be monitored for reliability. Modern HDC implementations have demonstrated efficient learning under resource constraints and strong robustness in sensing and recognition tasks \cite{rahimi2018efficient, imani2022neural}.

In manufacturing, HDC has been used for defect detection and quality monitoring where edge efficiency and intrinsic interpretability are requirements rather than afterthoughts \cite{chen2023brain, hoang2024hierarchical}. Graph-aware HDC formulations have also been explored, showing that relational structure can be represented through hypervector composition rather than through learned message passing \cite{poduval2022graphd, vashishth2019composition}. However, much of the manufacturing-focused HDC literature targets streaming signals or image pipelines where graphs represent sensor topology or hierarchical organization \cite{hoang2024hierarchical}. Per-sample PSP prediction imposes a different encoding problem: nodes are heterogeneous scalars with measurement-pipeline-dependent statistics, and dependencies are directed from process to structure to property. This setting requires both (i) composition rules that preserve directed dependency and (ii) node encoders that calibrate heterogeneous scalars before composition, which are not addressed by fixed random encoders in classical HDC.

PSP-HDC builds on these threads by combining a domain-defined directed PSP graph with graph-aligned hypervector composition and a trainable scalar-to-hypervector encoder that calibrates parameter-specific statistics prior to composition. This combination targets the per-sample PSP regime where interaction learning is underdetermined, while maintaining attribution pathways and efficient prototype-based inference.

\section{Research Methodology}\label{sec_methodology}
This section defines the PSP-HDC pipeline with rigorous notation. Let the dataset be
$\mathcal{D}=\{(\mathbf{x}_i,y_i)\}_{i=1}^{n}$, where $\mathbf{x}_i\in\mathbb{R}^{P}$ is the vector of $P$ scalar process and structure parameters for sample $i$, and $y_i\in\{1,\dots,C\}$ is its property class label. PSP-HDC maps each $\mathbf{x}_i$ to a $D$-dimensional hypervector $\mathbf{h}_i\in\mathbb{R}^{D}$, constructs class prototypes $\{\mathbf{m}_c\}_{c=1}^{C}$, and predicts by prototype retrieval using cosine similarity. The pipeline consists of four components. HDC primitives and prototype inference (Section~\ref{subsec_hdc_basics}). A trainable scalar-to-hypervector encoder that interfaces heterogeneous scalars with a fixed random hyperdimensional basis (Section~\ref{subsec_adaptive_encoding}). A directed PSP graph that constrains composition through graph-aligned binding and bundling (Section~\ref{subsec_psp_graph}). Intrinsic explainability and learning diagnostics via attribution and Memory Alignment and Separation (Sections~\ref{subsec_attribution} and~\ref{subsec_mas}).

\begin{figure*}[htbp]
  \centering
  \includegraphics[width=\linewidth]{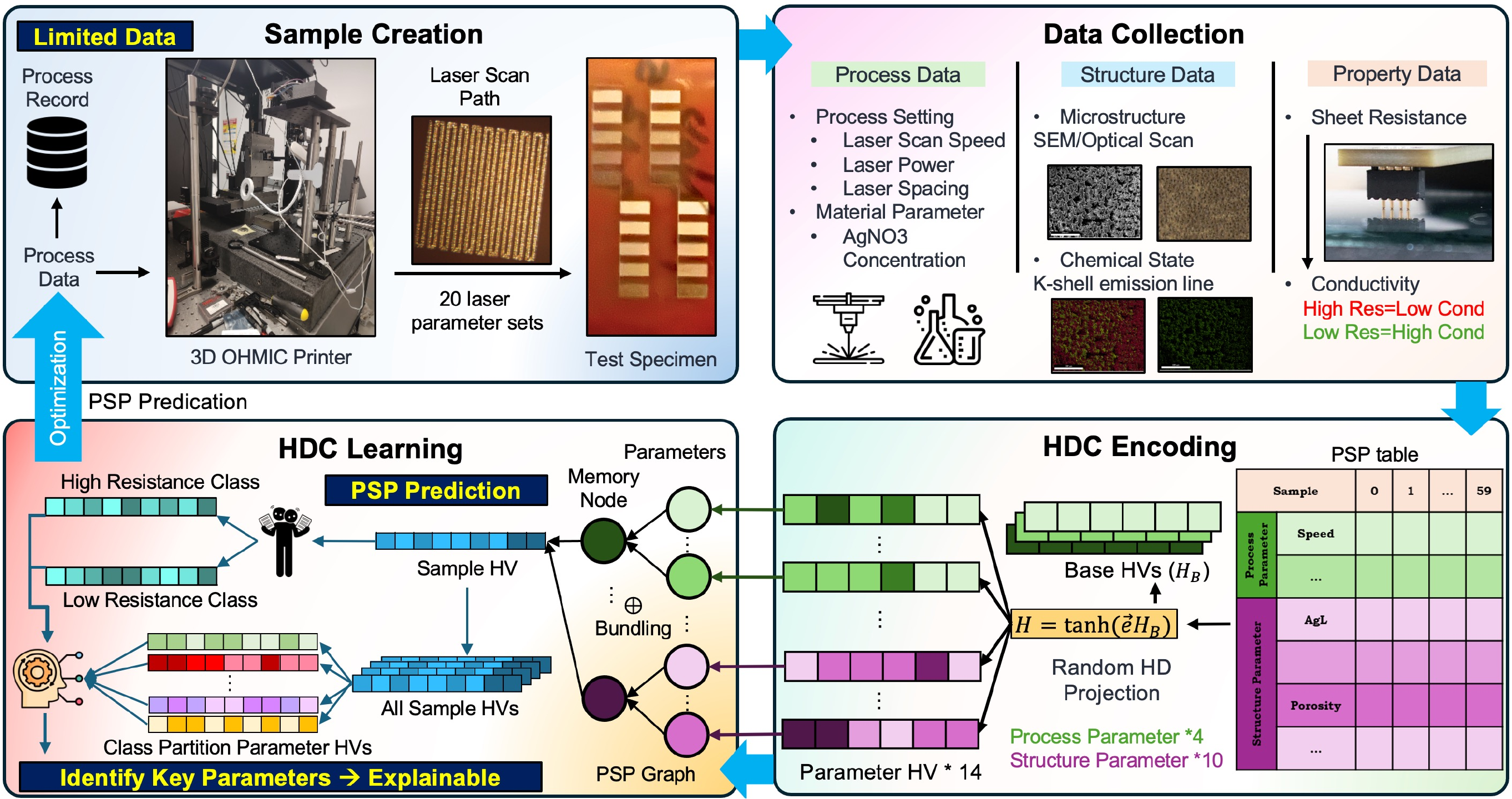}
    \caption{End-to-end PSP-HDC workflow for the multiphoton photoreduction dataset studied in this paper. Samples are fabricated under multiple process settings and fully characterized to obtain per-sample process parameters, structure descriptors, and property measurements. Each scalar parameter is encoded into a hypervector using a fixed random basis and trainable parameter-specific embeddings. Parameter hypervectors are composed along a directed PSP graph to form sample hypervectors and class prototype memories for similarity-based prediction. The same graph-consistent memories support parameter, group, and within-group attributions.}
  \label{fig:overview}
\end{figure*}

\subsection{HDC Basics}\label{subsec_hdc_basics}
Hyperdimensional computing represents entities as high-dimensional vectors and performs inference through compositional operators and associative-memory retrieval. PSP-HDC uses real-valued hypervectors in $\mathbb{R}^{D}$ with $D\in[10^{3},10^{4}]$, which yields near-orthogonality for random vectors and enables robust similarity-based retrieval.

\subsubsection{Hypervectors and normalization}
A hypervector is a column vector $\mathbf{h}\in\mathbb{R}^{D}$. We use the normalization operator
\begin{equation}
\operatorname{norm}(\mathbf{v})=
\begin{cases}
\mathbf{v}/\|\mathbf{v}\|_2, & \|\mathbf{v}\|_2>0\\
\mathbf{0}, & \|\mathbf{v}\|_2=0
\end{cases}
\label{eq_norm_def}
\end{equation}
where $\|\cdot\|_2$ is the Euclidean norm.

\subsubsection{Fixed random projection encoder}\label{subsubsec_random_projection}
A standard HDC encoder for a real-valued vector $\mathbf{z}\in\mathbb{R}^{p}$ is fixed random projection through a random basis matrix $\mathbf{B}\in\mathbb{R}^{p\times D}$ whose rows are fixed random base hypervectors \cite{thomas2021theoretical}. The encoded hypervector is
\begin{equation}
\mathbf{h}=\phi\!\left(\mathbf{z}^{\top}\mathbf{B}\right),
\qquad
\phi(t)=\tanh(t)\ \text{applied elementwise}
\label{eq_rp_encoder}
\end{equation}
The nonlinearity bounds the projection, yielding $\mathbf{h}\in(-1,1)^{D}$. This encoder is computationally lightweight but non-adaptive. For heterogeneous scalar PSP tables, it provides no mechanism to calibrate variable-specific scale and noise. PSP-HDC retains the fixed random basis but introduces trainable, parameter-specific embeddings that adapt each scalar node before projection (Section~\ref{subsec_adaptive_encoding}).

\subsubsection{Compositional operators}\label{subsubsec_core_ops}
PSP-HDC uses two operators on $\mathbb{R}^{D}$.
Binding is the Hadamard product
\begin{equation}
\mathbf{a}\odot\mathbf{b}=
\big[a_1b_1,\dots,a_Db_D\big]^{\top}
\label{eq_binding_def}
\end{equation}
Bundling is vector addition, optionally followed by normalization. For a finite index set $\mathcal{S}$ and hypervectors $\{\mathbf{u}_s\}_{s\in\mathcal{S}}$,
\begin{equation}
\begin{aligned}
\operatorname{bundle}\!\left(\{\mathbf{u}_s\}_{s\in\mathcal{S}}\right)=\sum_{s\in\mathcal{S}}\mathbf{u}_s\\
\operatorname{nbundle}\!\left(\{\mathbf{u}_s\}_{s\in\mathcal{S}}\right)=\operatorname{norm}\!\left(\sum_{s\in\mathcal{S}}\mathbf{u}_s\right)
\end{aligned}
\label{eq_bundling_def}
\end{equation}
In PSP-HDC, binding encodes directed dependencies, while bundling aggregates evidence within nodes and groups. These operations implement the PSP graph constraints because only graph-permitted bindings are constructed.

\subsubsection{Similarity and associative-memory inference}\label{subsubsec_similarity}
We measure similarity using cosine similarity
\begin{equation}
\operatorname{sim}(\mathbf{h}_1,\mathbf{h}_2)=
\frac{\mathbf{h}_1^{\top}\mathbf{h}_2}{\|\mathbf{h}_1\|_2\ \|\mathbf{h}_2\|_2}
\label{eq_cos_sim}
\end{equation}
If $\mathbf{h}_1$ and $\mathbf{h}_2$ are unit-normalized, $\operatorname{sim}(\mathbf{h}_1,\mathbf{h}_2)=\mathbf{h}_1^{\top}\mathbf{h}_2$.

\subsubsection{Prototype memory formation}\label{subsubsec_prototypes}
Let $\mathcal{T}\subseteq\{1,\dots,n\}$ be the training index set. For each class $c\in\{1,\dots,C\}$, define the class index set
\begin{equation}
\mathcal{T}_c=\{i\in\mathcal{T}\mid y_i=c\}
\label{eq_class_index_set}
\end{equation}
Given sample hypervectors $\{\mathbf{h}_i\}_{i\in\mathcal{T}}$, the class prototype is the normalized bundled sum
\begin{equation}
\mathbf{m}_c=\operatorname{norm}\!\left(\sum_{i\in\mathcal{T}_c}\mathbf{h}_i\right)\in\mathbb{R}^{D}
\label{eq_class_prototype}
\end{equation}
Prediction for a query $\mathbf{h}$ is prototype retrieval
\begin{equation}
\hat{y}=\arg\max_{c\in\{1,\dots,C\}}\operatorname{sim}(\mathbf{h},\mathbf{m}_c)
\label{eq_retrieval_rule}
\end{equation}

Some HDC pipelines apply prototype retraining by moving prototypes toward misclassified samples and away from confusable classes. If used, each update should be followed by normalization, for example
\begin{equation}
\mathbf{m}_{y}\leftarrow \operatorname{norm}(\mathbf{m}_{y}+\eta\,\mathbf{h}),
\qquad
\mathbf{m}_{\hat{y}}\leftarrow \operatorname{norm}(\mathbf{m}_{\hat{y}}-\eta\,\mathbf{h})
\label{eq_hdc_retrain}
\end{equation}
with $\eta>0$. In our setting, this step is not used because small class counts can cause prototype drift and degrade regime-holdout generalization.

\subsection{PSP Graph for PSP-HDC}\label{subsec_psp_graph}
PSP-HDC enforces process-to-structure-to-property topology through a directed PSP graph used as an internal structural prior. We define $K$ parameter groups and a partition of parameter indices $\{1,\dots,P\}$ into disjoint sets $\{\mathcal{S}_k\}_{k=1}^{K}$, where $\mathcal{S}_k$ contains the indices of parameters assigned to group $k$ and $\cup_{k=1}^{K}\mathcal{S}_k=\{1,\dots,P\}$.

For sample $i$, let $\mathbf{h}_{i,j}\in\mathbb{R}^{D}$ denote the hypervector encoding of scalar parameter $j$. The group hypervector is
\begin{equation}
\mathbf{n}_{i,k}=\operatorname{norm}\!\left(\sum_{j\in\mathcal{S}_k}\mathbf{h}_{i,j}\right)\in\mathbb{R}^{D}
\label{eq_group_hv}
\end{equation}

Directed dependencies between groups are specified by a directed graph on group indices, $\mathcal{G}=(\{1,\dots,K\},\mathcal{E})$, where $(u,k)\in\mathcal{E}$ indicates a directed edge from group $u$ to group $k$. Let the predecessor set of group $k$ be
\begin{equation}
\operatorname{pred}(k)=\{u\in\{1,\dots,K\}\mid (u,k)\in\mathcal{E}\}
\label{eq_pred_set}
\end{equation}
We encode directed dependencies by binding each group hypervector with its predecessors
\begin{equation}
\tilde{\mathbf{n}}_{i,k}=
\mathbf{n}_{i,k}\odot
\Bigg(\ \bigodot_{u\in\operatorname{pred}(k)}\mathbf{n}_{i,u}\ \Bigg)
\label{eq_graph_binding}
\end{equation}
where $\bigodot$ denotes repeated Hadamard product and the empty product is omitted, so if $\operatorname{pred}(k)$ is empty then $\tilde{\mathbf{n}}_{i,k}=\mathbf{n}_{i,k}$. This construction is the graph constraint, interaction terms exist only along edges in $\mathcal{E}$. The sample hypervector is then
\begin{equation}
\mathbf{h}_i=\operatorname{norm}\!\left(\sum_{k=1}^{K}\tilde{\mathbf{n}}_{i,k}\right)\in\mathbb{R}^{D}
\label{eq_sample_hv}
\end{equation}
The prototypes in Eq.~\eqref{eq_class_prototype} are built from $\{\mathbf{h}_i\}$, hence the PSP graph governs both inference and the geometry of class memories.

\begin{figure}[t]
  \centering
  \includegraphics[width=0.98\linewidth]{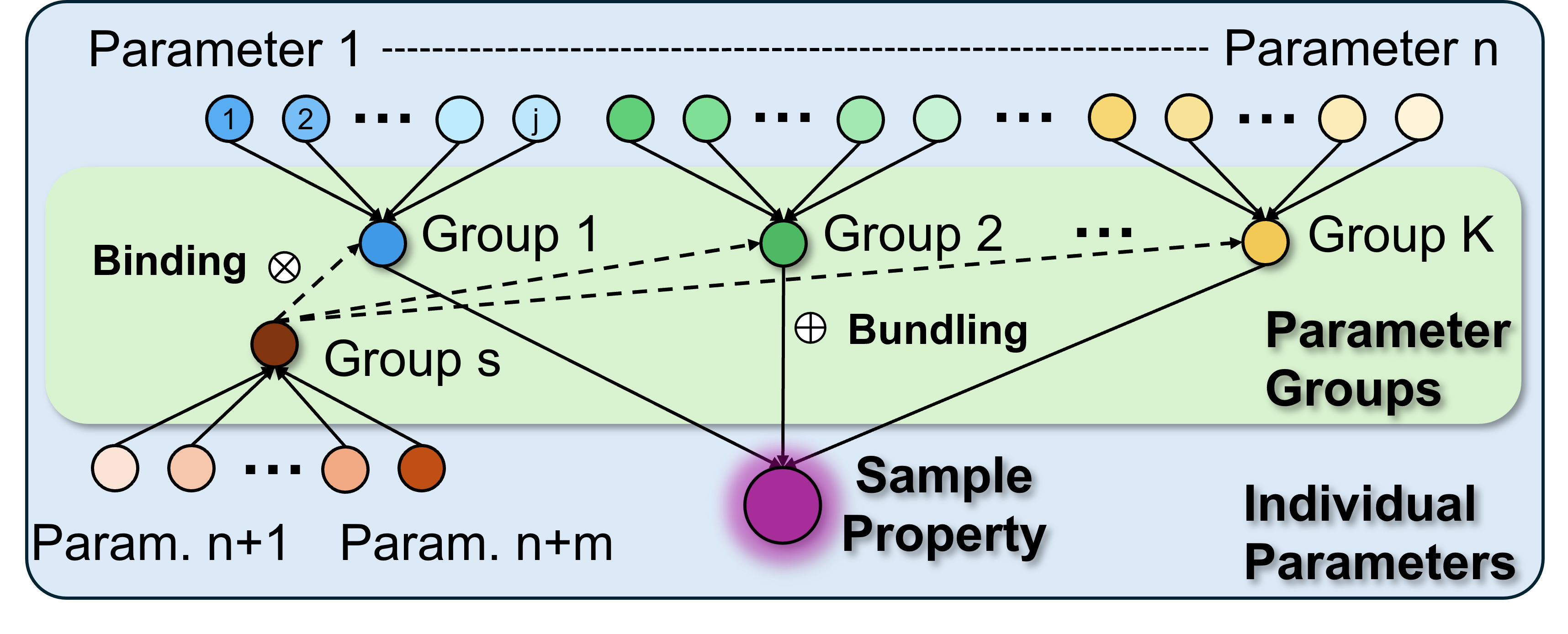}
  \caption{Illustrative PSP graph instantiated by bundling and graph-aligned binding. Scalars are bundled into group hypervectors, directed edges are encoded by binding group hypervectors along the specified dependencies, and the resulting group evidence is bundled into a sample hypervector for prototype learning.}
  \label{fig_graph_example}
\end{figure}

\subsection{Trainable Scalar-to-Hypervector encoding}\label{subsec_adaptive_encoding}
PSP graphs are only effective if their nodes are encoded consistently despite heterogeneous scales and noise. PSP-HDC therefore introduces a trainable scalar-to-hypervector encoder that keeps a fixed random basis but learns parameter-specific embeddings that calibrate each scalar node before graph composition.

Let $d\ll D$ be the embedding dimension. Let $\mathbf{B}\in\mathbb{R}^{d\times D}$ be a fixed random basis matrix whose rows are $\ell_2$ normalized. Each scalar parameter $j\in\{1,\dots,P\}$ has a trainable embedding vector $\mathbf{e}_j\in\mathbb{R}^{d}$. Only $\{\mathbf{e}_j\}_{j=1}^{P}$ are trained, so the encoder has $Pd$ learnable degrees of freedom.

\subsubsection{Scalar normalization}
For each parameter $j$, compute training-set extrema
\begin{equation}
x^{\min}_{j}=\min_{i\in\mathcal{T}} x_{i,j},
\qquad
x^{\max}_{j}=\max_{i\in\mathcal{T}} x_{i,j}
\label{eq_train_minmax}
\end{equation}

For any sample $i$, define the scaled scalar
\begin{equation}
\hat{x}_{i,j}=\frac{x_{i,j}-x^{\min}_{j}}{x^{\max}_{j}-x^{\min}_{j}+\varepsilon},
\qquad
\tilde{x}_{i,j}=2\,\operatorname{clip}_{[0,1]}(\hat{x}_{i,j})-1
\label{eq_signed_scaling}
\end{equation}
where $\varepsilon>0$ prevents division by zero and $\operatorname{clip}_{[0,1]}(t)=\min\{1,\max\{0,t\}\}$.

\subsubsection{Parameter hypervector encoding}
The parameter hypervector for scalar $x_{i,j}$ is
\begin{equation}
\mathbf{h}_{i,j}=
\tanh\!\left((\tilde{x}_{i,j}\mathbf{e}_{j})^{\top}\mathbf{B}\right)\in\mathbb{R}^{D}
\label{eq_param_hv}
\end{equation}

The trainable embedding $\mathbf{e}_j$ calibrates the node representation for parameter $j$ while the fixed basis $\mathbf{B}$ preserves the efficiency and stability of random hyperdimensional projection.

\begin{figure*}[htbp]
  \centering
  \includegraphics[width=\textwidth]{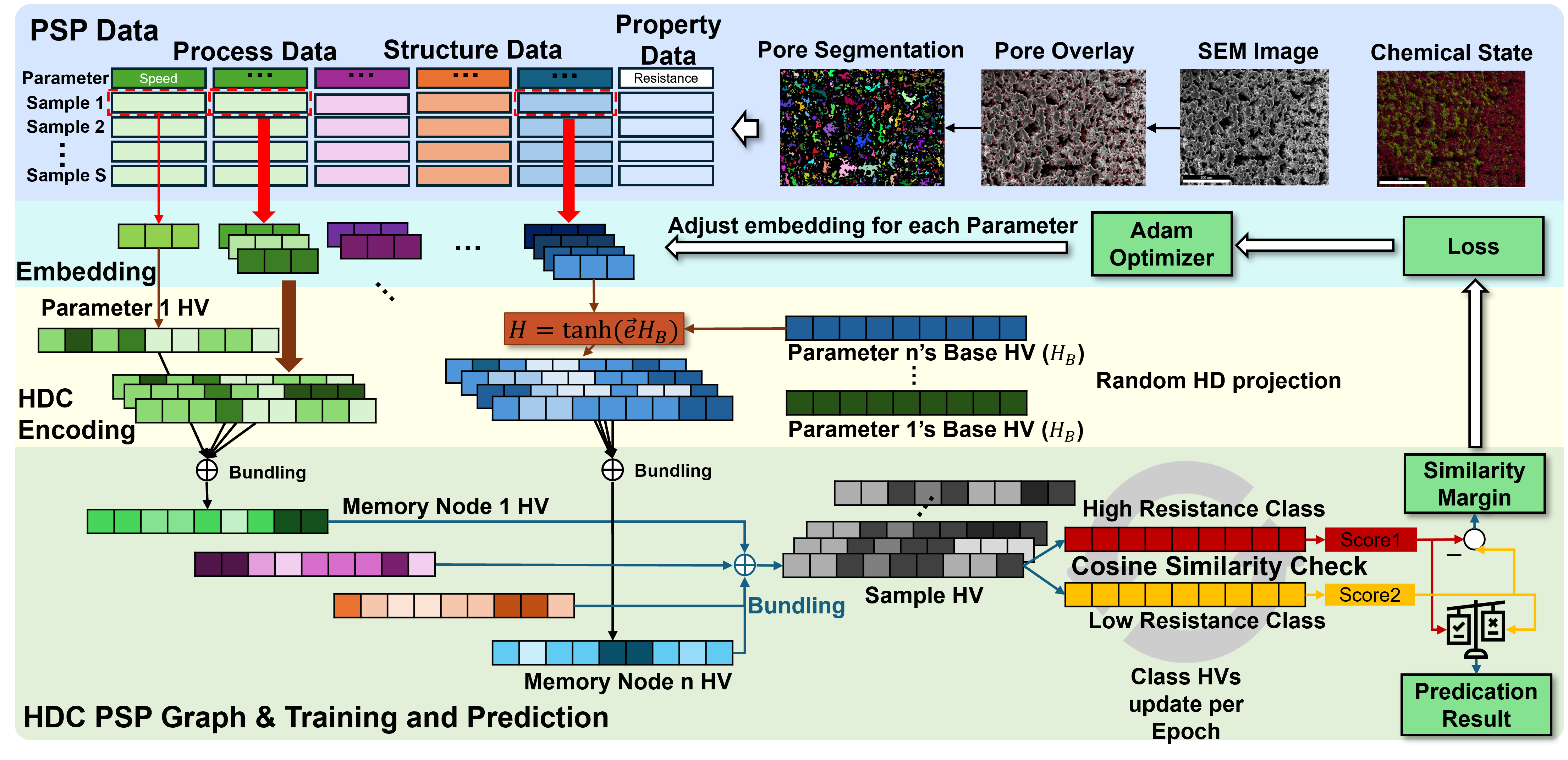}
  \caption{PSP-HDC training and inference. Each scalar $x_{i,j}$ is scaled to $\tilde{x}_{i,j}\in[-1,1]$, mapped to a $d$-dimensional parameter-specific embedding $\tilde{x}_{i,j}\mathbf{e}_j$, and projected to a $D$-dimensional hypervector via the fixed basis $\mathbf{B}$ as in Eq.~\eqref{eq_param_hv}. Parameter hypervectors are bundled into groups and composed along the directed PSP graph to obtain $\mathbf{h}_i$. Class prototypes $\{\mathbf{m}_c\}$ are formed by bundling training sample hypervectors within each class. Prediction uses cosine-similarity retrieval against prototypes, and the cross-entropy loss updates $\{\mathbf{e}_j\}$ while keeping $\mathbf{B}$ fixed.}
  \label{fig:hdc_workflow}
\end{figure*}

\subsubsection{Loss and optimization}\label{subsubsec_training}
At each epoch, compute sample hypervectors $\{\mathbf{h}_i\}_{i\in\mathcal{T}}$ using Eqs.~\eqref{eq_param_hv}–\eqref{eq_sample_hv}, then compute prototypes $\{\mathbf{m}_c\}_{c=1}^{C}$ by Eq.~\eqref{eq_class_prototype}. Define similarity logits for each training sample
\begin{equation}
z_{i,c}=\operatorname{sim}(\mathbf{h}_i,\mathbf{m}_c),
\qquad i\in\mathcal{T},\ c\in\{1,\dots,C\}
\label{eq_logits}
\end{equation}

The predicted class probabilities are
\begin{equation}
p_{i,c}=\frac{\exp(z_{i,c})}{\sum_{m=1}^{C}\exp(z_{i,m})}
\label{eq_softmax}
\end{equation}

The training objective is cross-entropy
\begin{equation}
\mathcal{L}=
-\frac{1}{|\mathcal{T}|}\sum_{i\in\mathcal{T}}\log p_{i,y_i}
\label{eq_ce_loss}
\end{equation}

The gradient of $\mathcal{L}$ is backpropagated through the scalar encoder, graph composition, and prototype construction to update $\{\mathbf{e}_j\}_{j=1}^{P}$ using Adam, while $\mathbf{B}$ remains fixed. Because prototypes are recomputed from the current embeddings, the optimization explicitly sharpens the geometry of class memories under the PSP graph constraints.

\subsection{Attribution-Based Explainability}\label{subsec_attribution}
PSP-HDC explanations are derived from the same graph-consistent memories used for inference. We compute class-partitioned component prototypes at parameter and group granularity. For each class $c$ and parameter $j$,
\begin{equation}
\mathbf{m}^{\mathrm{par}}_{c,j}=\operatorname{norm}\!\left(\sum_{i\in\mathcal{T}_c}\mathbf{h}_{i,j}\right)
\label{eq_param_memory}
\end{equation}

For each class $c$ and group $k$,
\begin{equation}
\mathbf{m}^{\mathrm{grp}}_{c,k}=\operatorname{norm}\!\left(\sum_{i\in\mathcal{T}_c}\mathbf{n}_{i,k}\right)
\label{eq_group_memory}
\end{equation}
where $\mathbf{n}_{i,k}$ is defined in Eq.~\eqref{eq_group_hv}. The property prototype is $\mathbf{m}_c$ in Eq.~\eqref{eq_class_prototype}.

Let $\mathbf{m}^{\mathrm{comp}}_{c,g}$ denote a generic component prototype, either $\mathbf{m}^{\mathrm{par}}_{c,j}$ for some parameter $j$ or $\mathbf{m}^{\mathrm{grp}}_{c,k}$ for some group $k$. For each $(c,g)$, define its affinity to each property prototype
\begin{equation}
a_{c,g,\ell}=\operatorname{sim}(\mathbf{m}^{\mathrm{comp}}_{c,g},\mathbf{m}_{\ell}),
\qquad \ell\in\{1,\dots,C\}
\label{eq_comp_affinity}
\end{equation}

Convert affinities to a class-affinity distribution using softmax with temperature $\beta>0$
\begin{equation}
\pi_{c,g,\ell}=
\frac{\exp(\beta a_{c,g,\ell})}{\sum_{m=1}^{C}\exp(\beta a_{c,g,m})}
\label{eq_comp_softmax}
\end{equation}

The class-consistent evidence of component $g$ for class $c$ is $q_{c,g}=\pi_{c,g,c}\in(0,1)$. Attribution weights over components are obtained by normalization across $g$
\begin{equation}
\alpha_{c,g}=\frac{q_{c,g}}{\sum_{h=1}^{G}q_{c,h}},
\qquad \sum_{g=1}^{G}\alpha_{c,g}=1
\label{eq_attrib_weights}
\end{equation}
where $G$ is the number of components in the chosen explanation granularity.

Within-group attribution is computed by restricting to parameters inside a group $k$. For $j\in\mathcal{S}_k$, define affinities
\begin{equation}
a^{(k)}_{c,j,\ell}=\operatorname{sim}(\mathbf{m}^{\mathrm{par}}_{c,j},\mathbf{m}^{\mathrm{grp}}_{\ell,k}),
\qquad \ell\in\{1,\dots,C\}
\label{eq_within_group_affinity}
\end{equation}

Then define $\pi^{(k)}_{c,j,\ell}$ by softmax as in Eq.~\eqref{eq_comp_softmax} with $a^{(k)}_{c,j,\ell}$, set $q^{(k)}_{c,j}=\pi^{(k)}_{c,j,c}$, and normalize within the group
\begin{equation}
\alpha^{(k)}_{c,j}=
\frac{q^{(k)}_{c,j}}{\sum_{m\in\mathcal{S}_k} q^{(k)}_{c,m}},
\qquad \sum_{j\in\mathcal{S}_k}\alpha^{(k)}_{c,j}=1
\label{eq_within_group_weights}
\end{equation}

This yields parameter, group, and within-group attributions aligned with the directed PSP topology.

\subsection{Memory Alignment and Separation}\label{subsec_mas}
Memory Alignment and Separation quantifies how learning sharpens component memories relative to property prototypes. For each class $c$ and component $g$, reuse the affinities $a_{c,g,\ell}$ in Eq.~\eqref{eq_comp_affinity} and the induced distribution $\pi_{c,g,\ell}$ in Eq.~\eqref{eq_comp_softmax}. Alignment is the certainty assigned to the correct class
\begin{equation}
A_{c,g}=\pi_{c,g,c}
\label{eq_mas_alignment}
\end{equation}

Separation is the margin between the correct-class certainty and the strongest competing certainty
\begin{equation}
S_{c,g}=\pi_{c,g,c}-\max_{\ell\in\{1,\dots,C\}\setminus\{c\}}\pi_{c,g,\ell}
\label{eq_mas_separation}
\end{equation}

Per-class ($c\in\{1,\dots,C\}.$) summaries are
\begin{equation}
\begin{aligned}
\mathrm{MA}_c &= \frac{1}{G}\sum_{g=1}^{G}A_{c,g}, \qquad
\mathrm{MS}_c = \frac{1}{G}\sum_{g=1}^{G}S_{c,g} \\
\end{aligned}
\label{eq_mas_summary}
\end{equation}

Increases in $\mathrm{MA}_c$ and $\mathrm{MS}_c$ indicate that class-partitioned component memories become more aligned with the correct property prototype and more separated from competing prototypes under the PSP graph constraints.

\section{Experiment Design} \label{sec: Experiment design}

This section describes how we fabricate specimens and construct the per-sample PSP table used for learning. The experiments were carried out on the 3D OHMIC research platform, but all specimens analyzed in this study were fabricated using its multiphoton photoreduction mode. As summarized in Fig.~1, we fabricate specimens under multiple process-parameter settings; for each build, the process settings and relevant records are logged to support subsequent analysis and reproducibility (Section~\ref{sec: Process data}). After fabrication and post-processing, we perform structure characterization using SEM and extract scalar structure parameters from the images (Section~\ref{sec: Structure data}). In addition, we acquire chemical-state/composition information using SEM-EDS (EDAX TEAM) and include the resulting measurements as structure parameters (Section~\ref{sec: Structure data}). Finally, we measure the electrical property of interest, the sheet resistance, and record it as the target property label/response for each sample (Section~\ref{sec:Property data}). All process parameters, structure parameters, and property measurements are collated into a unified PSP table with one row per sample, which serves as the input to the PSP-HDC framework described next.

\subsection{Process Data} \label{sec: Process data}
\begin{figure}[htbp]
  \centering
  \includegraphics[width=\linewidth]{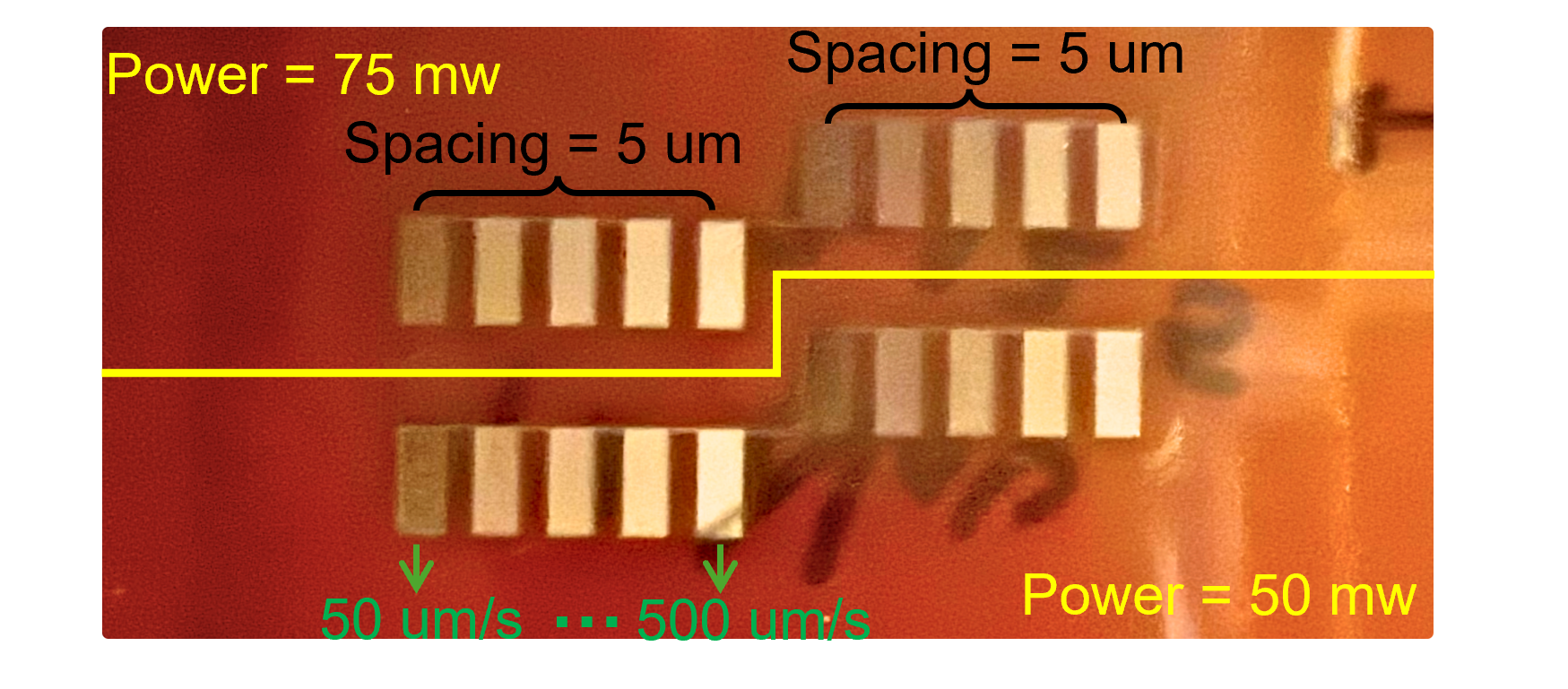}
  \caption{Representative fabricated sample array produced using the 3D OHMIC multiphoton photoreduction process. Two laser power settings ($P = 75$ and $50$~mW) and two hatch spacings ($h = 5$ and $10~\mu$m) define four blocks, and within each block the scan speed is varied across five levels ($v = 50$--$500~\mu$m/s). This layout spans the full $2 \times 2 \times 5 = 20$ machine-parameter combinations used for each AgNO$_3$ concentration group.}
  \label{fig:sample}
\end{figure}

We employ a full-factorial design over one material/chemistry factor and three machine-controlled factors (Table~\ref{tab:param_levels}), yielding $3 \times 2 \times 2 \times 5 = 60$ unique process conditions with one fabricated specimen per condition. AgNO$_3$ concentration is the material/chemistry factor, whereas laser power, hatch spacing, and scan speed are machine-controlled factors (Table~\ref{tab:param_levels}). Based on AgNO$_3$ concentration, the dataset is partitioned into three material groups (5\%, 10\%, and 15\%). Within each group, we exhaustively sweep the same set of machine-controlled parameters, yielding $2 \times 2 \times 5 = 20$ distinct machine-parameter combinations per concentration level (Table~\ref{tab:param_levels}). All specimens in this study were fabricated using the multiphoton photoreduction capability of the 3D OHMIC platform. The femtosecond laser was operated at 120~fs pulse duration, 80~MHz repetition rate, and 800~nm wavelength using a $10\times$ objective (NA = 0.25). Figure~\ref{fig:sample} shows representative fabricated samples from the resulting process matrix.
In conventional scan-based laser processes, these three variables are often collapsed into a single scalar parameter, the areal energy density (AED),
\begin{equation}
\mathrm{AED}=\frac{P}{v h}
\end{equation}
where AED has units of mJ/$\mu$m$^2$, $P$ is the laser power (mW), $v$ is the scan speed ($\mu$m/s), and $h$ is the hatch spacing ($\mu$m). Equivalently, $P/v$ is the linear energy input (mJ/$\mu$m) along a scan track, and dividing by $h$ distributes this energy over the scanned area, thereby reducing the effective dimensionality of the process space. However, this reduction is not well justified for the present study. The dataset analyzed here is generated by multiphoton photoreduction, and the process--structure relationship is governed by photoreduction kinetics and optical energy localization rather than by energy deposition alone. As a result, different combinations of $(P,v,h)$ can produce materially different reduction behavior even at similar AED. We therefore retain power, hatch spacing, and scan speed as independent process parameters instead of compressing them into a single surrogate, yielding a more physically grounded PSP prediction problem. Process data collection is therefore straightforward since we record these controlled settings for each fabricated sample.

\begin{table}[htbp]
\centering
\caption{Process parameter-levels and experimental design.}
\label{tab:param_levels}
\begin{tabular}{l l}
\hline
\textbf{Parameter} & \textbf{Levels} \\
\hline
AgNO$_3$ Concentration (\%) & 5, 10, 15 \\
Power (mW) & 50, 75 \\
Spacing ($\mu$m) & 5, 10 \\
Scan speed ($\mu$m/s) & 50, 100, 200, 300, 500 \\
\hline
Total samples & $3 \times 2 \times 2 \times 5 = 60$ \\
\hline
\end{tabular}
\end{table}

\subsection{Structure Data} \label{sec: Structure data}
Structure parameters are extracted from SEM-based characterization. All SEM micrographs used for morphology quantification are acquired at a fixed magnification (Fig.~\ref{fig:sem_5pct_speed_sweep}; scale bar $100~\mu\mathrm{m}$) under consistent imaging conditions, so pixel-based statistics are proportional to physical length/area; we report features in pixels for reproducibility and convert to physical units when required for interpretation. We group structure descriptors into (i) chemical-element parameters from SEM-EDS (EDAX TEAM) and (ii) pore-morphology parameters from SEM imaging combined with image processing.

\subsubsection{Chemical element (SEM-EDS, EDAX TEAM).}
\begin{figure}[htbp]
  \centering
  \includegraphics[width=\linewidth]{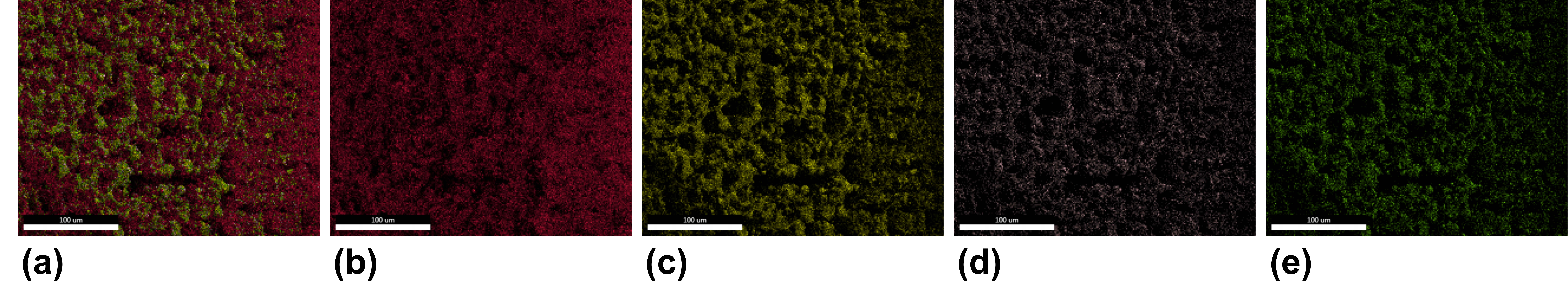} 
  \caption{SEM-EDS elemental maps shown as a single montage for Sample~0 (5\% AgNO$_3$, $P=50$~mW, $h=5~\mu\mathrm{m}$, $v=50~\mu\mathrm{m}/\mathrm{s}$). Panels (a)-(e) correspond to: (a) SEM image with all EDS channels overlaid, (b) Ag (Ag~L), (c) C (C~K), (d) N (N~K), and (e) O (O~K) distribution maps.}
  \label{fig:element}
\end{figure}

For each sample, SEM-EDS provides elemental overlay maps for Ag, C, N, and O (Fig.~\ref{fig:element}(b)-(e)) as well as an all-channel overlay (Fig.~\ref{fig:element}(a)), together with spectrum-based quantification using the eZAF workflow.

\subsubsection{Pore-related morphology (SEM + image processing).} 
\begin{figure}[htbp]
  \centering
  \includegraphics[width=\linewidth]{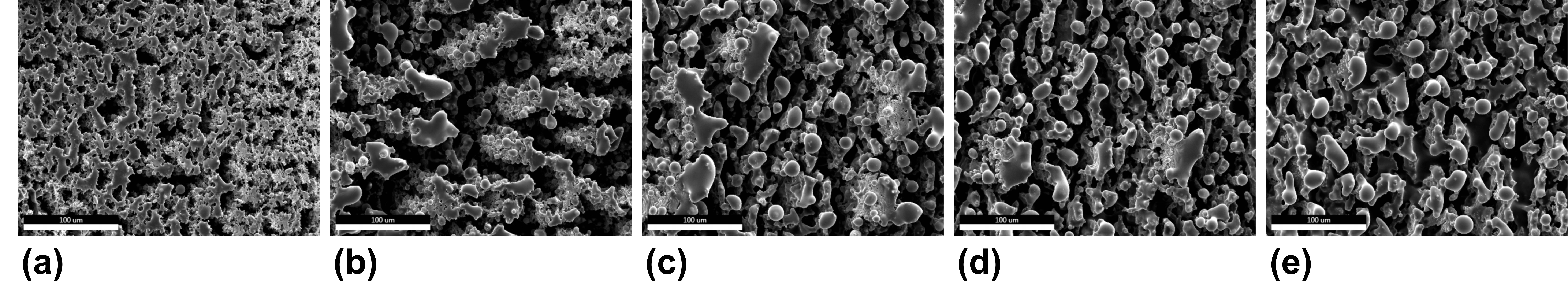} 
  \caption{SEM montage for the 5\% AgNO$_3$ group under a scan-speed sweep at fixed laser power and hatch spacing ($P=50$~mW, $h=5~\mu\mathrm{m}$). Panels (a)-(e) correspond to: (a) $v=50~\mu\mathrm{m}/\mathrm{s}$, (b) $v=100~\mu\mathrm{m}/\mathrm{s}$, (c) $v=200~\mu\mathrm{m}/\mathrm{s}$, (d) $v=300~\mu\mathrm{m}/\mathrm{s}$, and (e) $v=500~\mu\mathrm{m}/\mathrm{s}$. All images are shown at the same magnification; scale bar: $100~\mu\mathrm{m}$.}
  \label{fig:sem_5pct_speed_sweep}
\end{figure}

We posit that SEM-based structure characterization provides additional structural cues that are informative for predicting sheet resistance. Figure~\ref{fig:sem_5pct_speed_sweep} illustrates this with a simple example from the 5\% AgNO$_3$ group, where laser power and hatch spacing are fixed at $P=50$~mW and $h=5~\mu$m and only the scan speed is varied ($v=50$-$500~\mu$m/s). Under this setup, Sample~0 ($v=50~\mu$m/s) exhibits a finite sheet resistance of 17.02, whereas the remaining samples yield effectively insulating behavior (reported as $\infty$) and show more similar microstructural appearances to each other than to Sample~0. This qualitative contrast motivates incorporating structure-derived parameters into the PSP dataset. 

\begin{figure}[htbp]
  \centering
  \includegraphics[width=\linewidth]{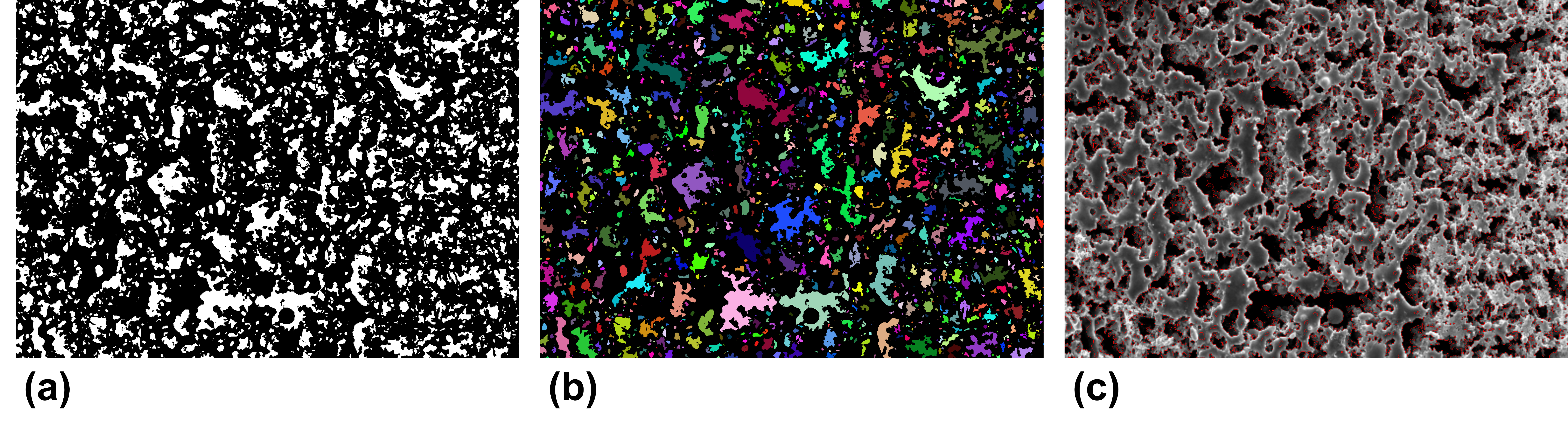} 
  \caption{Pore segmentation and verification from an SEM micrograph of Sample 0. Panels (a)-(c) correspond to: (a) binary pore mask obtained via Sauvola adaptive thresholding (dark regions identified as pores), (b) connected-component pore labels where colors denote distinct pores, and (c) overlay of segmented pore boundaries (red) on the original SEM image for visual verification.}
  \label{fig:pore_pipeline}
\end{figure}

Accordingly, pore statistics are computed from SEM micrographs via a standardized image-processing pipeline. We first apply Sauvola adaptive thresholding~\cite{sauvola2000adaptive} (implemented in \texttt{scikit-image}) to binarize the SEM image and obtain a pore mask (Fig.~\ref{fig:pore_pipeline}(a)), where dark regions are identified as pores; this example corresponds to Sample~0, whose original SEM micrograph is shown in Fig.~\ref{fig:sem_5pct_speed_sweep}(a). Connected pore regions are then labeled (each pore assigned a unique label/color) to enable pore-wise measurements and visualization (Fig.~\ref{fig:pore_pipeline}(b)). Based on these labeled pore regions, we compute pore-morphology parameters from the segmented pore phase~\cite{gostick2019porespy}. Finally, we overlay the segmented boundaries (red contours) on the original SEM image as a visual verification of the pore extraction (Fig.~\ref{fig:pore_pipeline}(c)). 

We compute 17 pore-morphology parameters, including porosity (area fraction of the pore phase), number of pores ($N_\mathrm{pore}$), pore-size statistics based on an equivalent pore radius $r$ (mean and standard deviation, in pixels), and pore-area statistics $A$ (mean, standard deviation, median, and upper-tail percentiles $p90$ and $p99$, in pixels$^2$). We report the area coefficient of variation ($\mathrm{CV}=\sigma_A/\mu_A$), the geometric mean and geometric standard deviation of pore area, and the fraction of total pore area contributed by the largest 1\% pores. To capture characteristic length scales beyond equivalent radius, we compute local-thickness (LT) diameter statistics (mean/median, in pixels) and chord-length parameters along the horizontal and vertical directions (mean chord lengths in pixels).

As part of data quality control, Samples 14 and 39 were excluded because incorrect scale calibration was identified in their SEM images. Compared with the remaining samples, these images have a larger scale and a smaller field of view, which could introduce inconsistency in pore-related morphological characterization. Therefore, all subsequent analyses were conducted using the remaining 58 samples.

\subsection{Property Data} \label{sec:Property data}
The property of interest is the sheet resistance $R_s$ of each fabricated sample, reported in $\Omega/sq$. Sheet resistance is a standard thin-film metric that characterizes the intrinsic in-plane electrical resistance of a conductive layer independent of its lateral dimensions. For a uniform film, the measured resistance between opposite sides scales with the length-to-width ratio rather than the absolute size; equivalently, $R_s$ can be interpreted as the resistance of a square patch of the film. This is well aligned with 3D OHMIC, where the functional outcome of interest is the lateral electrical transport of the printed conductive features/layers; therefore, $R_s$ provides a direct, geometry-normalized measure of electrical performance for comparing process conditions across the fabricated dataset.

The measurement of sheet resistance is achieved by using a collinear four-probe (four-point probe / Kelvin) method, in which a known current $I$ is driven through the two outer probes while the voltage drop $V$ is sensed between the two inner probes~\cite{smits1958fourpoint}. Because the voltage-sensing circuit draws negligible current, the measured $V$ is largely insensitive to probe/contact resistances compared to two-probe measurements~\cite{smits1958fourpoint}. The measured resistance is
\begin{equation}
R_{\mathrm{meas}}=\frac{V}{I}.
\end{equation}

For an effectively infinite sheet with equal probe spacing, the sheet resistance is given by~\cite{smits1958fourpoint}
\begin{equation}
R_s=\frac{\pi}{\ln 2}\,\frac{V}{I}\,C
\label{eq:4pp_rs}
\end{equation}
where $C$ is a dimensionless correction factor that accounts for finite sample geometry, edge effects, and thickness/spatial constraints; tabulated correction factors for practical geometries are available in~\cite{swartzendruber1964tn199}. In our measurements, the instrument reports $R_s$ directly in $\Omega/\mathrm{sq}$.

In some cases, the instrument returns an ``$\infty$'' reading, indicating that the sheet resistance exceeds the measurable range (right-censored observation). For visualization and numerical summaries that require finite values, we map ``$\infty$'' to the maximum finite sheet resistance observed within the corresponding AgNO$_3$ concentration group. Let $c\in\{5\%,10\%,15\%\}$ denote the concentration group and define
\begin{equation}
R_s^{\max}(c)=\max\{\,R_s \in c:\; R_s<\infty\,\}
\end{equation}

We then define the clipped value used only for plotting/summaries,
\begin{equation}
\tilde{R}_s=\min\bigl(R_s,\;R_s^{\max}(c)\bigr)
\label{eq:rs_clip}
\end{equation}
so that all out-of-range readings are displayed at the group-wise censoring level (e.g., $R_s^{\max}(5\%)=48{,}860~\Omega/\text{sq}$).

\subsection{Sample distribution}
Sheet resistance $R_s$ is highly skewed and spans multiple orders of magnitude across the 60 fabricated samples. We therefore summarize $R_s$ on a logarithmic scale (i.e., $\log_{10}(R_s)$), where a unit difference corresponds to a tenfold change. A subset of measurements is reported as ``$\infty$'' because the resistance exceeds the instrument range (effectively insulating). These out-of-range readings are clipped to the maximum finite sheet resistance within the corresponding 20-sample subset; this preprocessing can lead to repeated high-resistance values. Table~\ref{tab:rs_summary} reports the resulting dataset-level statistics.

Given the limited dataset size, we formulate property prediction as a binary classification problem. The split point $R_s^\star$ is selected via a distance-based criterion in log space: after sorting the finite $R_s$ values, we identify the largest adjacent gap in $\log_{10}(R_s)$, which lies between $123.8$ and $552.7~\Omega/\mathrm{sq}$. We place the threshold at the midpoint of this gap (equivalently, the geometric mean), $R_s^\star=\sqrt{123.8\times 552.7}\approx 2.62\times10^{2}~\Omega/\mathrm{sq}$, and round to $R_s^\star=260~\Omega/\mathrm{sq}$ in the experiments. This yields 25 samples in the high-resistance class and 35 samples in the low-resistance class. With additional fabricated samples, the property space could be discretized into more resistance levels to support multi-class modeling.

\begin{table}[t]
\centering
\caption{Summary statistics of sheet resistance $R_s$ for the 3D OHMIC PSP dataset.}
\label{tab:rs_summary}
\begin{tabular}{l c}
\toprule
\textbf{Statistic} & \textbf{Value} \\
\midrule
Total samples ($N$) & 60 \\
Out-of-range readings reported as ``$\infty$'' & 16 (26.7\%) \\
Finite readings used for numerical summaries & 44 (73.3\%) \\
\midrule
$R_s$ (finite) min-max [$\Omega/\mathrm{sq}$] & 0.6157 - $4.886\times10^{4}$ \\
$R_s$ (finite) median [Q1, Q3] [$\Omega/\mathrm{sq}$] & 4.02 [1.557, 117.35] \\
$R_s$ (finite) 5-95\% range [$\Omega/\mathrm{sq}$] & 0.695 - $1.153\times10^{4}$ \\
$\log_{10}(R_s)$ (finite) median [Q1, Q3] & 0.604 [0.192, 2.069] \\
\bottomrule
\end{tabular}
\end{table}

\subsection{PSP graph for 3D OHMIC} \label{sec: PSP graph for 3D OHMIC}
\begin{figure}[htbp]
  \centering
  \includegraphics[width=\linewidth]{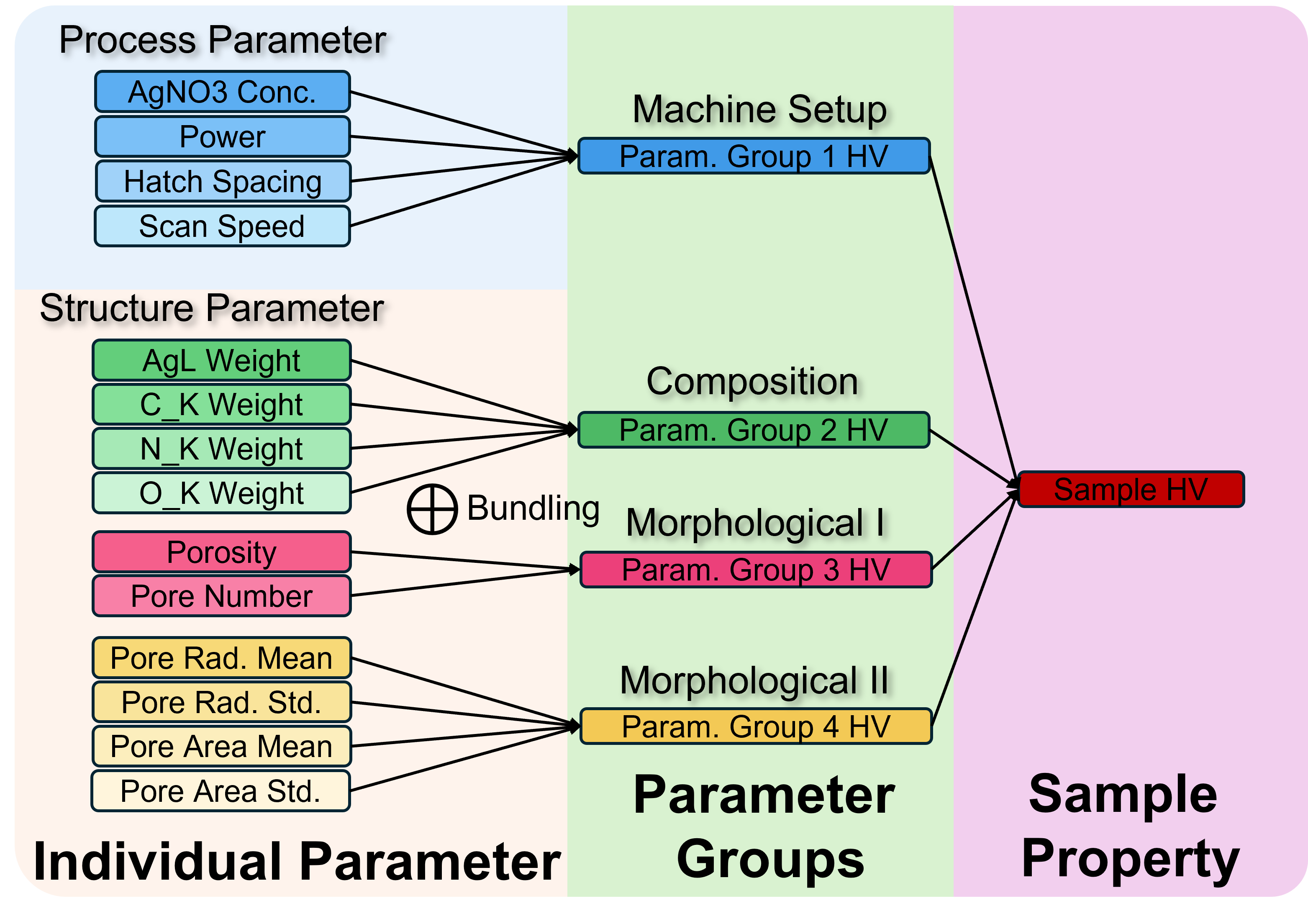}
  \caption{Each parameter is represented as a parameter-level node (encoded as a hypervector) and is assigned to a physics-motivated parameter group (dashed boxes) for group-wise composition. The Process Parameter group bundles four process parameters. The Structure Parameter is organized into three parameter groups (parameter groups): Composition (chemical-state parameters from SEM-EDS elemental weights) and two pore-morphology groups (Morphological I and Morphological II). Parameter group hypervectors are formed by bundling the contained parameter hypervectors and are subsequently bundled to produce the sample hypervector for prototype memory learning and attribution.}
  \label{fig:graph}
\end{figure}

As illustrated in Fig.~\ref{fig:graph}, we instantiate the PSP graph for the 3D~OHMIC workflow by grouping the available PSP parameters into semantically coherent parameter groups based on physical meaning. The first layer includes process parameters (four machine parameters: AgNO$_3$ concentration, laser power, hatch spacing, and scan speed) and structure parameters (four elemental composition and available pore-morphology parameters). We further organize these structure parameters into three physics-motivated groups: Composition, Morphological I, and Morphological II.

\section{Experimental Results} \label{sec: Results}
This section validates PSP-HDC on the 3D OHMIC PSP table and connects predictive performance to internal model behavior. The central empirical fact is that the sheet-resistance response is interaction dominated under small, heterogeneous PSP records. Univariate process or structure trends provide only weak separation and cannot support reliable regime prediction or mechanistic ranking. PSP-HDC addresses this failure mode by enforcing a directed PSP topology during representation and prototype formation, then exposing decisions through graph-consistent memories. We report results in five steps. We first quantify why the task is ill posed for feature-vector rules using preliminary PSP dependence analysis (Section~\ref{sec: Preliminary Analysis}). We then assess hyperparameter sensitivity of the adaptive encoder that interfaces heterogeneous scalars with hyperdimensional graph composition (Section~\ref{sec: PSP-HDC sensitivity analysis}). Next, we benchmark predictive performance under both random splits and regime-holdout process folds to test robustness and cross-regime transfer (Section~\ref{sec: Benchmark comparison}). We then analyze how training reshapes prototype geometry and sample alignment, including MAS and sample-level movement in hypervector space (Section~\ref{sec: PSP-HDC learning dynamics}). Finally, we report intrinsic explainability through parameter, group, and within-group attributions derived from class-partitioned memories (Section~\ref{sec: PSP-HDC Explainability}). 

\subsection{Preliminary Analysis} \label{sec: Preliminary Analysis}
We first examine empirical dependencies along the process-structure-property chain to expose the source of difficulty in 3D OHMIC PSP prediction. Across the dataset, no single process or structure parameter is a sufficient discriminator for sheet-resistance regime. Marginal trends are weak, non-monotonic, or heavily overlapping between classes. Different samples can share similar values for an individual parameter yet yield resistance values separated by orders of magnitude. This is the signature of a coupled PSP mapping that requires joint modeling rather than per-parameter rules.

We examine direct process to property relations using the four process parameters in 3D OHMIC, AgNO$_3$ concentration, laser power, hatch spacing, and scan speed. All four show strong class overlap when viewed independently, and the apparent effect of any one parameter depends on the others. This interaction structure motivates a compositional representation that encodes joint process effects instead of relying on marginal trends.

Figure~\ref{fig:rs_overview}(a) shows sheet resistance on a $\log_{10}$ scale across three concentrations. The 5\% formulation contains many effectively insulating outcomes reported as $\infty$ and clipped for visualization, producing a dense cluster at the top of the plotted range. The 15\% formulation concentrates more heavily in the low-resistance regime with fewer extreme values, while 10\% exhibits the broadest spread and spans multiple orders of magnitude. Concentration shifts the distribution and the frequency of extreme outcomes, but it does not uniquely determine $R_s$.

\begin{figure}[htbp]
  \centering
  \includegraphics[width=1\linewidth]{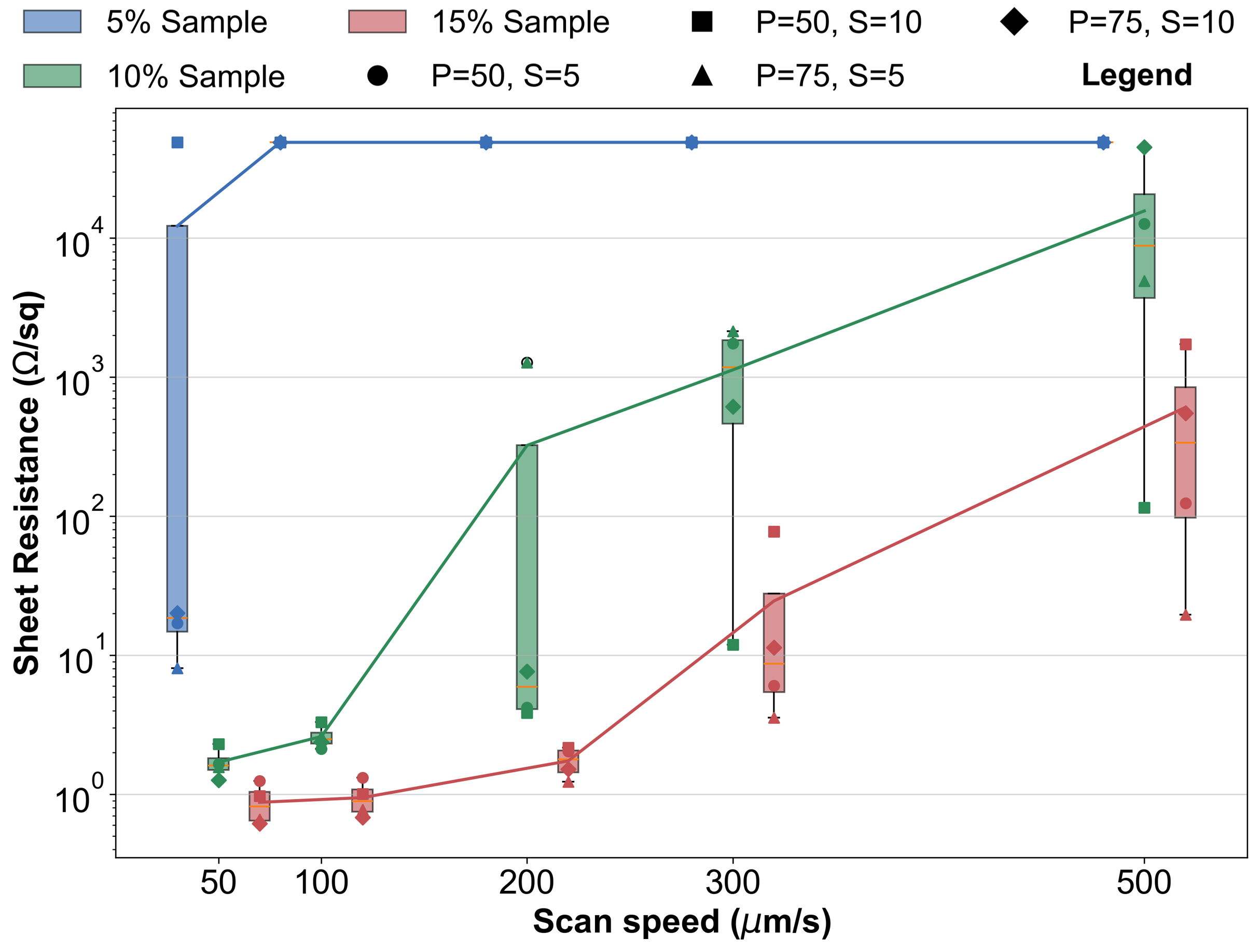}
  \caption{Sheet resistance $R_s$ distributions and marginal trends versus AgNO$_3$ concentration in the 3D OHMIC dataset. The vertical axis is logarithmic and one decade corresponds to a tenfold change. Markers denote individual samples. Numeric labels indicate scan speed. Box plots report the median, interquartile range, and whiskers extending to $1.5\times$IQR. Repeated maximum-resistance values reflect visualization clipping of out-of-range and $\infty$ readings to the largest finite value within the corresponding subset.}
  \label{fig:rs_overview0}
\end{figure}

\begin{figure*}[t]
  \centering
  \includegraphics[width=1\linewidth]{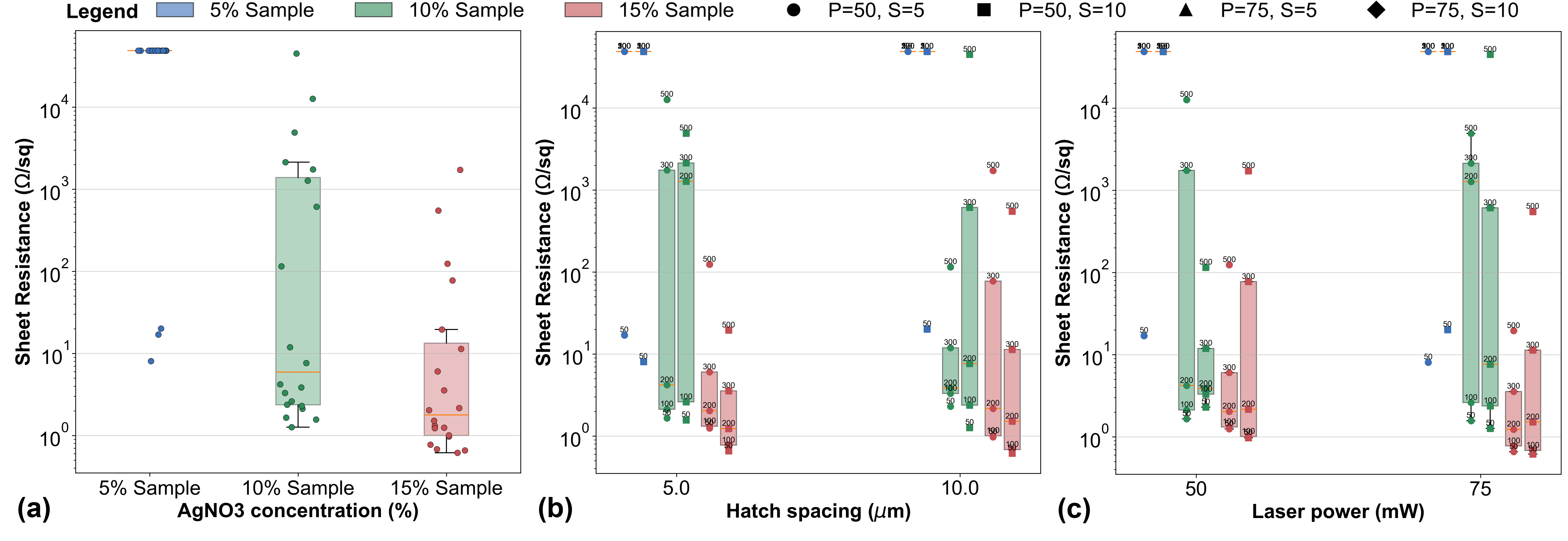}
  \caption{Sheet resistance distributions and marginal trends across key process variables in the 3D OHMIC dataset. The vertical axis is logarithmic and one decade corresponds to a tenfold change. Panel (a) shows $R_s$ versus scan speed with box plots summarizing all measurements at each speed and concentration-conditioned median trends indicated by colored lines. Panels (b) and (c) show $R_s$ versus hatch spacing and laser power, with subgroup distributions within each concentration. Markers denote individual samples. Numeric labels indicate scan speed. Box plots report the median, interquartile range, and whiskers extending to $1.5\times$IQR. Repeated maximum-resistance values reflect visualization clipping of out-of-range and $\infty$ readings to the largest finite value within the corresponding subset.}
  \label{fig:rs_overview}
\end{figure*}

Figure~\ref{fig:rs_overview0} reveals a pronounced speed dependence. $R_s$ generally increases as scan speed increases, consistent with reduced delivered energy per unit length at higher speeds. Low-speed settings more frequently achieve low sheet resistance, whereas high-speed settings produce larger $R_s$ and greater dispersion, including many high-resistance realizations. The pattern is strongest for the 10\% and 15\% formulations. The 5\% formulation remains dominated by insulating or clipped outcomes across speeds, which demonstrates an interaction between scan speed and formulation.

Figure~\ref{fig:rs_overview}(b) shows that hatch spacing does not deterministically set $R_s$. Substantial dispersion persists even when power is fixed, and the annotated scan-speed values highlight strong interaction effects. In the 10\% formulation, increasing spacing from 5 to 10~$\mu$m tends to reduce the spread of $R_s$ and suppress extreme high-resistance outcomes, suggesting a more consistent response in this concentration. In the 15\% formulation, the 10~$\mu$m setting increases variability and introduces high-$R_s$ outliers compared with 5~$\mu$m, particularly under $P=50$~mW. This concentration-dependent reversal rules out hatch spacing as a standalone control variable.

Figure~\ref{fig:rs_overview}(c) shows that power effects are entangled with spacing and scan speed. For the 10\% formulation, both $P=50$ and $75$~mW yield order-of-magnitude variation in $R_s$, and increasing power does not guarantee monotonic improvement. High-resistance outcomes remain present, especially under faster scans. For the 15\% formulation, low-$R_s$ outcomes are more prevalent overall, yet widening spacing and higher scan speeds again coincide with increased dispersion and high-$R_s$ excursions, even at $P=75$~mW. These results demonstrate that no single process parameter acts as a reliable knob for $R_s$. The property response is governed by coupled, non-additive interactions among formulation, power, spacing, and speed.

Overall, the process-only plots reveal only coarse tendencies. Higher AgNO$_3$ concentration and lower scan speed increase the frequency of low-resistance outcomes. These tendencies are not sufficient for process design because large within-setting variability persists at fixed concentration and speed, and extremely low scan speed is often incompatible with throughput objectives. This motivates PSP predictors that integrate process and structure evidence and explicitly model coupled effects.

\subsubsection{Structure to property}
\begin{figure*}[t]
  \centering
  \includegraphics[width=1\linewidth]{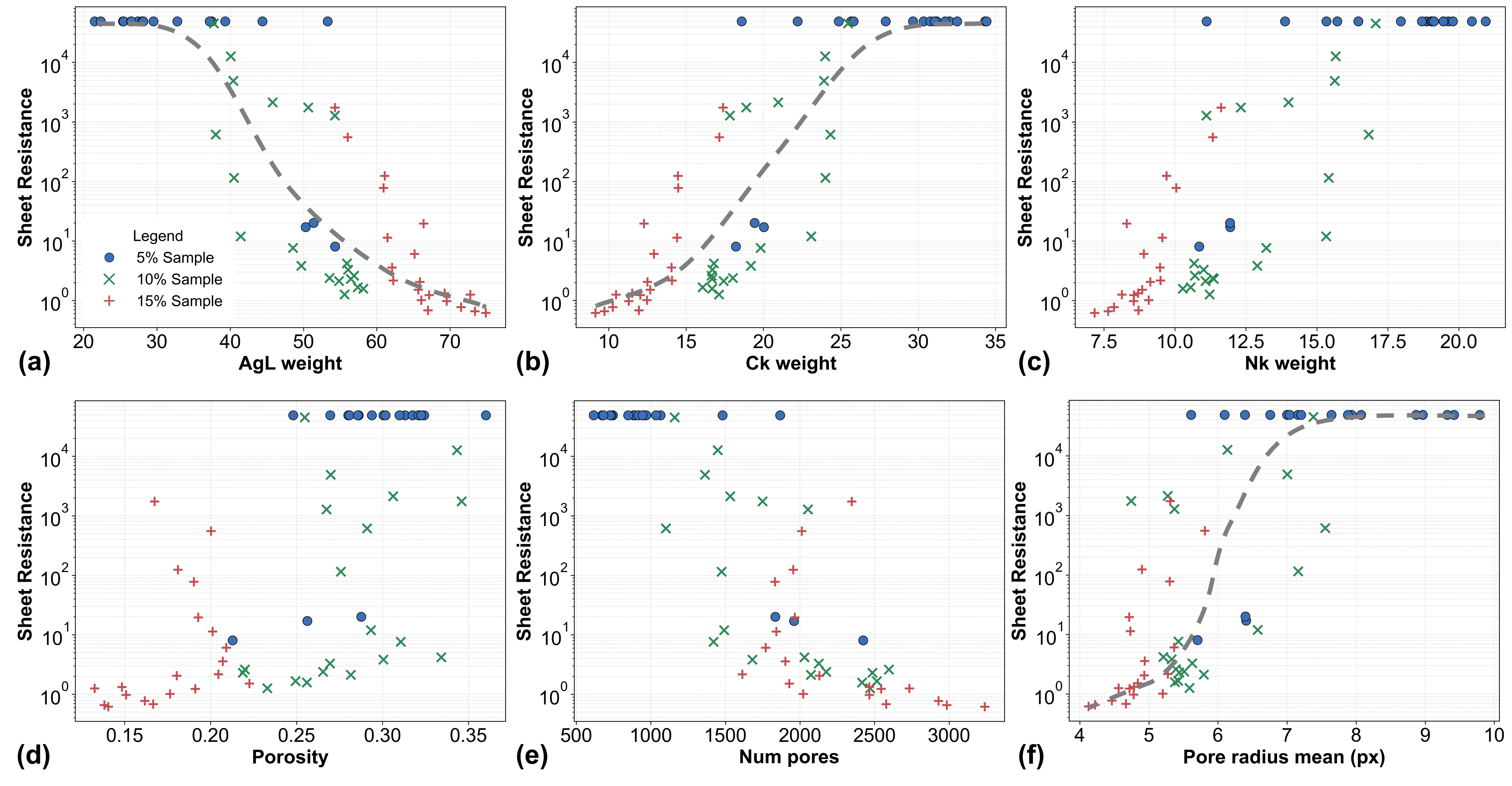}
  \caption{Structure to property relations for representative composition and pore-morphology descriptors. Sheet resistance $R_s$ is shown on a logarithmic scale versus AgL weight, Ck weight, Nk weight, porosity, number of pores, and mean pore radius in pixels. Markers denote AgNO$_3$ concentration groups. Gray dashed curves provide guides to the eye for coarse tendencies. Substantial scatter and multiple outliers persist, and several parameters exhibit weak or non-monotonic marginal trends. Overall, the figure indicates that $R_s$ is driven by coupled structural effects, motivating group-wise modeling rather than single-parameter rules.}
  \label{fig:structure2property}
\end{figure*}

Figure~\ref{fig:structure2property} examines representative structure descriptors versus $R_s$ on a logarithmic scale with markers indicating the three concentration groups. Several parameters show coarse threshold-like tendencies in isolation, including composition-related weights and mean pore-radius statistics. These marginal trends do not support reliable prediction. Samples with similar values of a single descriptor still span orders of magnitude in $R_s$ across broad regions of the structure space. For several morphology parameters such as porosity and pore count, the relation to $R_s$ is weak or non-monotonic when aggregated over the full dataset, with outliers that violate any one-parameter rule.

These observations indicate that $R_s$ is governed by coupled structure effects rather than by a single descriptor. Similar porosity can arise from distinct pore configurations such as a few large pores versus many small pores, which plausibly induce different conductive pathways. This motivates grouped modeling of morphology and composition descriptors and joint inference, which PSP-HDC implements through PSP graph composition and group-level representations.

\subsection{PSP-HDC Sensitivity Analysis}\label{sec: PSP-HDC sensitivity analysis}
We next assess sensitivity of PSP-HDC to hyperparameters in the trainable scalar-to-hypervector encoder, which controls how heterogeneous scalars populate PSP graph nodes. All settings are evaluated using 1000 random 80\% and 20\% train and test splits. We report mean test accuracy across splits. We fix the hypervector dimension at $D=5000$ and sweep embedding dimension $d\in\{4,8,16,32,64\}$, learning rate $\eta\in\{10^{-4},5\times10^{-3},10^{-3},5\times10^{-2},10^{-2}\}$, and training horizon $E\in\{100,200,300,400,500\}$ epochs for a total of 125 configurations.

For each pair $(d,\eta)$, we track mean accuracy as a function of $E$ and summarize by the best achievable accuracy within the tested horizons
\begin{equation}
\begin{aligned}
A^{\star}(d,\eta)=\max_{E\in\{100,200,300,400,500\}} \bar{a}(d,\eta,E),\\
E^{\star}(d,\eta)=\arg\max_{E\in\{100,200,300,400,500\}} \bar{a}(d,\eta,E),
\end{aligned}
\end{equation}
where $\bar{a}(d,\eta,E)$ is the mean test accuracy across 1000 splits at epoch budget $E$.

\begin{figure}[htbp]
  \centering
  \includegraphics[width=\linewidth]{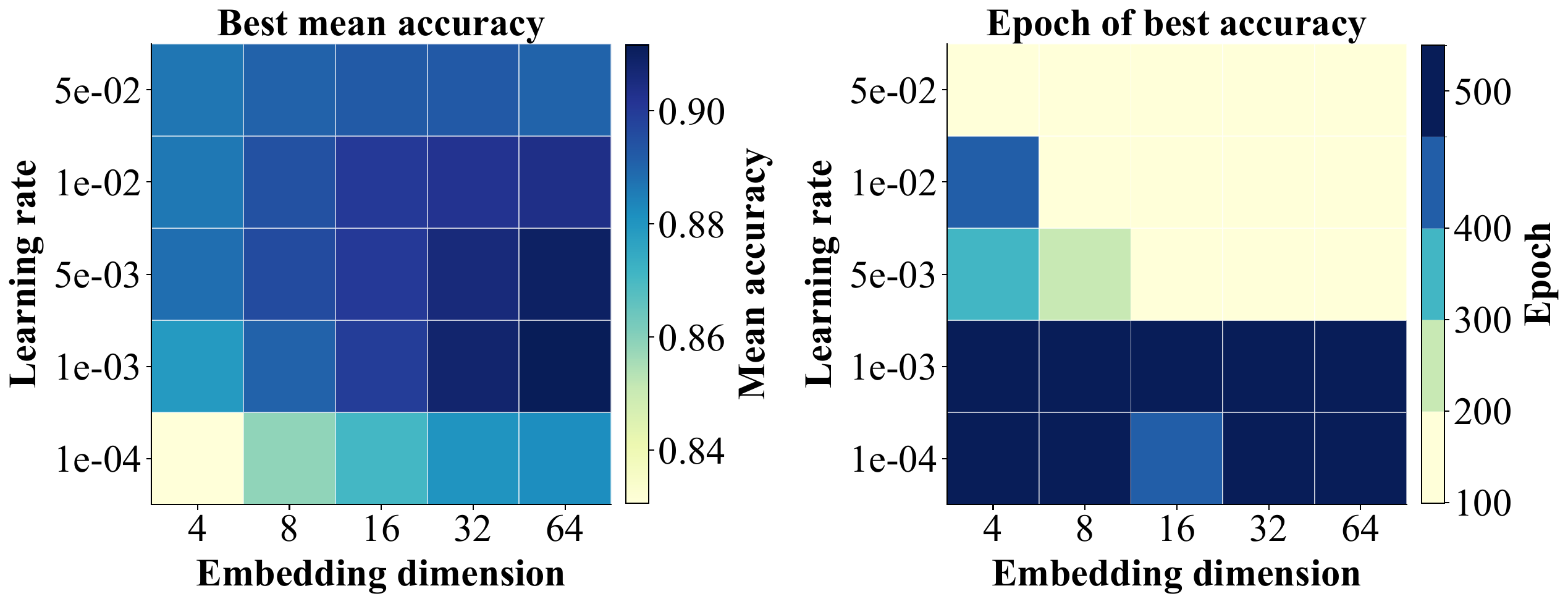}
  \caption{Sensitivity of the adaptive scalar encoder with fixed hypervector dimension $D=5000$. The left heatmap shows $A^{\star}(d,\eta)$, the best mean accuracy for each $(d,\eta)$ over the tested epoch budgets. The right heatmap shows $E^{\star}(d,\eta)$, the epoch budget at which that best accuracy is achieved.}
  \label{fig:stage1_edim_lr}
\end{figure}

Figure~\ref{fig:stage1_edim_lr} reports $A^{\star}(d,\eta)$ and $E^{\star}(d,\eta)$. Two conclusions are immediate. First, high accuracy is not confined to a single narrow hyperparameter point. A broad region of $(d,\eta)$ achieves near-peak accuracy, which indicates that PSP-HDC is not a brittle configuration-dependent method. Second, performance saturates for moderate embedding dimensions. Increasing $d$ beyond 32 yields only marginal gains, which supports the design premise that parameter-specific calibration can be achieved with a compact embedding while preserving HDC efficiency.

Table~\ref{tab:stage1_results} lists the top 10 configurations. We select $d=32$, $\eta=10^{-3}$, and $E=300$ as the default for subsequent experiments because it lies in the high-accuracy plateau and offers a favorable cost-to-accuracy trade-off. Differences between this sweep and later benchmark numbers are driven by split variability and are expected for $n=60$.

\begin{table}[t]
\centering
\caption{Top 10 hyperparameter combinations for sensitivity analysis of the adaptive scalar encoder with fixed hypervector dimension $D=5000$.}
\label{tab:stage1_results}
\begin{tabular}{c c c c c}
\toprule
Combination & $d$ & Learning rate & Epochs & Accuracy (mean $\pm$ std) \\
\midrule
114 & 64 & 0.001 & 500 & $0.911583 \pm 0.077248$ \\
105 & 64 & 0.005 & 100 & $0.909500 \pm 0.079434$ \\
113 & 64 & 0.001 & 400 & $0.908917 \pm 0.081409$ \\
 89 & 32 & 0.001 & 500 & $0.907833 \pm 0.081341$ \\
 80 & 32 & 0.005 & 100 & $0.905583 \pm 0.076889$ \\
112 & 64 & 0.001 & 300 & $0.905083 \pm 0.081806$ \\
 88 & 32 & 0.001 & 400 & $0.904417 \pm 0.080337$ \\
120 & 64 & 0.010 & 100 & $0.903583 \pm 0.078544$ \\
 95 & 32 & 0.010 & 100 & $0.902083 \pm 0.078721$ \\
 81 & 32 & 0.005 & 200 & $0.901833 \pm 0.076572$ \\
\bottomrule
\end{tabular}
\end{table}

\subsection{Benchmark Comparison} \label{sec: Benchmark comparison}
Explainability is only meaningful when the underlying predictions are accurate and robust. We therefore benchmark PSP-HDC against representative baselines trained on the same measured PSP table without synthetic augmentation. The baselines include SVM, logistic regression, RNN, CNN, Na\"ive Bayes, $k$-nearest neighbors, random forest trees, and Gaussian process classification. We evaluate under two protocols. Random splits quantify robustness to sampling variability through repeated 80\% and 20\% train and test partitions. Process folds test cross-regime generalization by holding out an entire region of the process space during training.

\subsubsection{Random split test}
Table~\ref{tab:random_split_benchmark} reports mean and standard deviation of accuracy, precision, recall, and F1 score across 1000 random splits. All methods exhibit non-trivial split-to-split variation, which is expected for $n=60$ with overlapping PSP regimes and borderline samples. PSP-HDC achieves the highest mean accuracy and the highest mean F1 score, with a competitive dispersion. The strongest classical baselines cluster in the 0.875 to 0.885 mean accuracy range, while CNN, random forests, and GPC are lower and more variable. The RNN baseline collapses in this setting, consistent with the mismatch between sequential architectures and small static PSP tables.

The performance gap is not marginal. PSP-HDC improves mean accuracy by roughly 2.5 to 3.5 percentage points over the best non-HDC baselines and improves F1 similarly, while maintaining stable variability. This establishes a rigorous performance foundation for the mechanistic interpretation of prototype memories and attributions.

\begin{table*}[htbp]
\centering
\caption{Benchmark performance under 1000 random 80\% and 20\% train and test splits. Values are mean $\pm$ standard deviation across splits. Best values per metric are typeset in boldface.}
\label{tab:random_split_benchmark}
\small
\setlength{\tabcolsep}{4pt}
\begin{tabular}{lcccc}
\toprule
Model & Accuracy ($mean\!\pm\!std$) & Precision ($mean\!\pm\!std$) & Recall ($mean\!\pm\!std$) & F1-score ($mean\!\pm\!std$) \\
\midrule
SVM     & $0.8845 \pm 0.0823$ & $0.8994 \pm 0.0798$ & $0.8751 \pm 0.0880$ & $0.8774 \pm 0.0880$ \\
LR      & $0.8787 \pm 0.0823$ & $0.8927 \pm 0.0804$ & $0.8706 \pm 0.0872$ & $0.8718 \pm 0.0877$ \\
RNN     & $0.5545 \pm 0.1967$ & $0.5511 \pm 0.2149$ & $0.5436 \pm 0.1949$ & $0.5346 \pm 0.2004$ \\
CNN     & $0.8384 \pm 0.1136$ & $0.8660 \pm 0.1100$ & $0.8316 \pm 0.1107$ & $0.8253 \pm 0.1276$ \\
NB      & $0.8817 \pm 0.0818$ & $0.8945 \pm 0.0799$ & $0.8760 \pm 0.0853$ & $0.8759 \pm 0.0860$ \\
KNN     & $0.8753 \pm 0.0806$ & $0.8974 \pm 0.0746$ & $0.8607 \pm 0.0895$ & $0.8648 \pm 0.0900$ \\
RFT     & $0.8373 \pm 0.0928$ & $0.8580 \pm 0.0922$ & $0.8261 \pm 0.0989$ & $0.8254 \pm 0.1024$ \\
GPC     & $0.8326 \pm 0.0875$ & $0.8504 \pm 0.0879$ & $0.8248 \pm 0.0900$ & $0.8232 \pm 0.0936$ \\
PSP-HDC & $\mathbf{0.9103 \pm 0.0769}$ & $\mathbf{0.9241 \pm 0.0718}$ & $\mathbf{0.9007 \pm 0.0838}$ & $\mathbf{0.9044 \pm 0.0823}$ \\
\bottomrule
\end{tabular}
\end{table*}

\subsubsection{Process fold test}
Random splits do not test the practical goal in early-stage process development, namely prediction in previously unobserved process regimes. We therefore evaluate process-regime generalization using process folds. Each fold holds out all samples from one regime, trains on the remaining regimes, and evaluates exclusively on the held-out regime. We consider three regime definitions aligned with the 3D OHMIC design space, AgNO$_3$ concentration, laser power, and hatch spacing. This protocol is substantially harder than random splits because the test set occupies a distinct region of the process space, and for binary regime variables it reduces the training set to approximately half the samples.

Table~\ref{tab:process_fold_benchmark} reports held-out accuracy per regime. PSP-HDC achieves the strongest or tied-best performance on every fold and yields the highest mean accuracy across folds. Gains are most pronounced in the most distribution-shifted cases, including the 5\% concentration and 50\,mW power holdouts, where several baselines degrade sharply. In contrast, PSP-HDC remains stable, which indicates that its directed PSP graph and prototype-memory retrieval capture transferable coupled structure rather than regime-specific coincidences. This cross-regime reliability is the operative requirement for accelerating process exploration in 3D OHMIC.

\begin{table*}[htbp]
\centering
\caption{Process-fold generalization accuracy. Each row holds out all samples from one process regime, trains on the remaining regimes, and evaluates on the held-out regime. Best performance per row is typeset in boldface.}
\label{tab:process_fold_benchmark}
\small
\setlength{\tabcolsep}{4pt}
\begin{tabular}{lccccccccc}
\toprule
Held-out regime & SVM & LR & RNN & CNN & NB & KNN & RFT & GPC & PSP-HDC \\
\midrule
AgNO$_3$ 5\%   & 0.5263 & 0.7368 & 0.1579 & 0.8947 & 0.8947 & 0.8421 & 0.8947 & 0.7895 & $\mathbf{0.9474}$ \\
AgNO$_3$ 10\%  & 0.7895 & 0.7368 & 0.6316 & 0.7368 & 0.7368 & 0.7895 & 0.6842 & 0.6842 & $\mathbf{0.8421}$ \\
AgNO$_3$ 15\%  & 0.6500 & $\mathbf{0.9000}$ & $\mathbf{0.9000}$ & $\mathbf{0.9000}$ & $\mathbf{0.9000}$ & $\mathbf{0.9000}$ & $\mathbf{0.9000}$ & $\mathbf{0.9000}$ & $\mathbf{0.9000}$ \\
Laser power 50\,mW & 0.6000 & 0.8667 & 0.6000 & 0.8667 & 0.8667 & 0.9000 & 0.8000 & 0.8333 & $\mathbf{0.9333}$ \\
Laser power 75\,mW & 0.4286 & 0.8571 & 0.6071 & 0.6786 & $\mathbf{0.8929}$ & 0.7857 & 0.8214 & 0.8214 & $\mathbf{0.8929}$ \\
Hatch spacing 5\,$\mu$m  & 0.7931 & $\mathbf{0.8966}$ & 0.5862 & $\mathbf{0.8966}$ & 0.8621 & 0.7931 & 0.8276 & 0.7931 & $\mathbf{0.8966}$ \\
Hatch spacing 10\,$\mu$m & 0.7586 & 0.7586 & 0.6897 & 0.7931 & 0.7586 & $\mathbf{0.8621}$ & 0.7586 & 0.7586 & $\mathbf{0.8621}$ \\
\midrule
Mean & 0.6494 & 0.8218 & 0.5961 & 0.8238 & 0.8445 & 0.8389 & 0.8124 & 0.7972 & $\mathbf{0.8963}$ \\
\bottomrule
\end{tabular}
\end{table*}

\subsection{PSP-HDC Learning Dynamics} \label{sec: PSP-HDC learning dynamics}
We now analyze how PSP-HDC achieves its accuracy gains by inspecting training dynamics beyond final metrics. We use the process fold that holds out the AgNO$_3$ 10\% regime as a representative case. This regime exhibits a mixed distribution of sheet-resistance classes relative to the more polarized 5\% and 15\% regimes in Fig.~\ref{fig:rs_overview}(a). The resulting class balance yields stable prototypes for both classes and provides a stringent setting for probing representation learning. We analyze three aspects. Global optimization and prototype stability. Memory-level geometry quantified by MAS. Sample-level movement toward prototypes.

\subsubsection{Global learning behavior and prototype stability}
\begin{figure}[htbp]
  \centering
  \includegraphics[width=\linewidth]{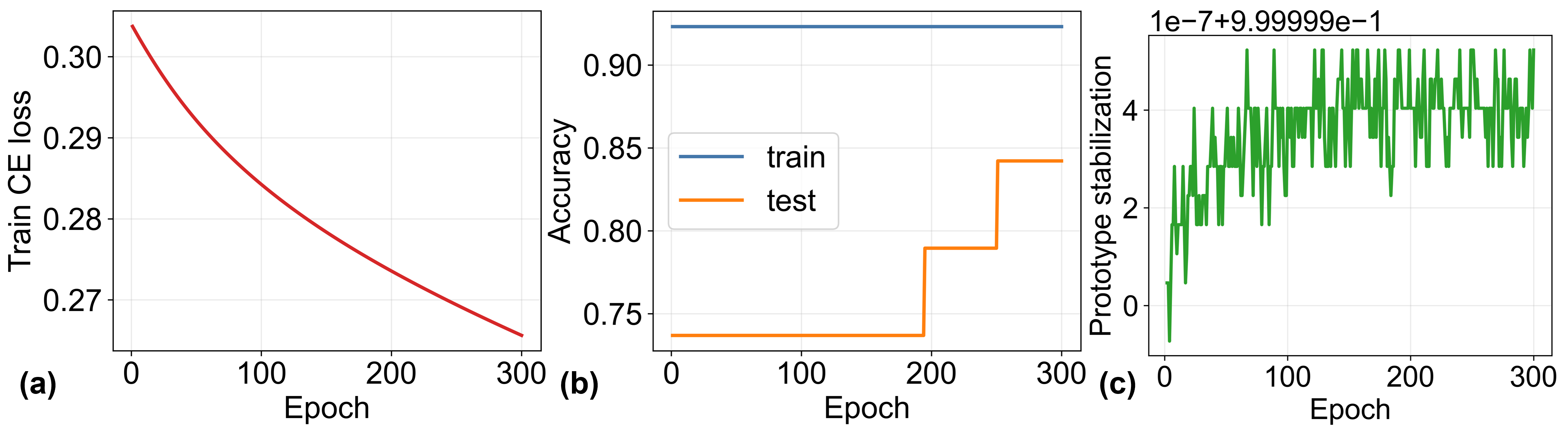}
  \caption{Global learning behavior and prototype evolution for PSP-HDC on the AgNO$_3$ 10\% holdout. Panel (a) shows training cross-entropy versus epoch. Panel (b) shows train and test accuracy versus epoch. Panel (c) shows prototype stabilization via $\bar{\rho}(t)$ in Eq.~\eqref{eq:proto_stability}.}
  \label{fig:dyn_global}
\end{figure}

Figure~\ref{fig:dyn_global}(a) shows the training cross-entropy for the adaptive encoder decreasing smoothly, indicating consistent improvement of similarity-based retrieval under the learned scalar embeddings. Figure~\ref{fig:dyn_global}(b) shows that training accuracy is high throughout, while test accuracy improves later in training, indicating that the encoder continues to refine class separation under the held-out regime. To quantify prototype stability, let $\mathbf{m}_c^{(t)}$ denote the class-$c$ property prototype at epoch $t$ computed from the current encoded sample hypervectors. We measure epoch-to-epoch prototype similarity by
\begin{equation}
\bar{\rho}(t)=\frac{1}{C}\sum_{c=1}^{C}\operatorname{sim}\!\left(\mathbf{m}_c^{(t)},\mathbf{m}_c^{(t-1)}\right)
\label{eq:proto_stability}
\end{equation}

Figure~\ref{fig:dyn_global}(c) shows that $\bar{\rho}(t)$ stabilizes after an initial transient, indicating that prototype drift is minimal after early epochs. Notably, the largest gains in test accuracy occur after prototypes have already stabilized, which implies that late-stage improvements are primarily driven by improved encoding and sample alignment rather than by continued prototype relocation. This is the critical distinction from fixed-encoder HDC pipelines where only prototype adjustment is available.

\subsubsection{Memory Alignment and Separation}
\begin{figure*}[htbp]
  \centering
  \includegraphics[width=\linewidth]{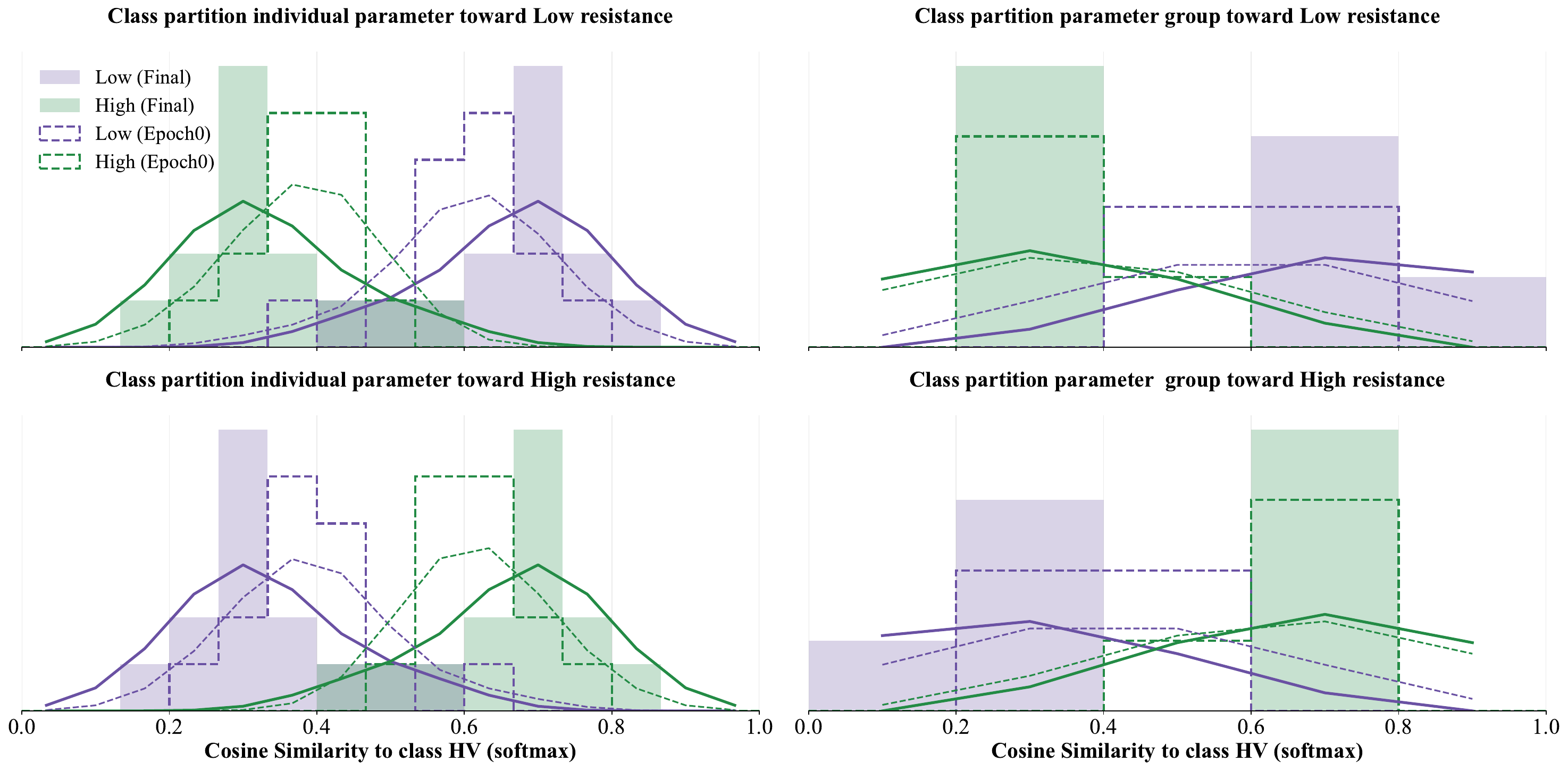}
  \caption{MAS distribution comparison for class-partitioned components before training and after convergence. The left column shows individual parameters and the right column shows parameter groups. The top row shows preference toward the low-resistance prototype and the bottom row shows preference toward the high-resistance prototype. Post-training distributions exhibit stronger separation between class partitions, indicating improved alignment with the correct prototype and reduced affinity to the competing prototype.}
  \label{fig:mas_certainty}
\end{figure*}

We next quantify how learning reshapes class-partitioned component memories at two granularities, individual parameters and parameter groups. Figure~\ref{fig:mas_certainty} compares MAS distributions at initialization and after convergence. At both granularities, training increases class-consistent preference and reduces confusion. Component memories in each class partition move toward their property prototype and away from the competing prototype.

Table~\ref{tab:mas_summary_results} reports per-class mean alignment and separation. Parameter-level alignment increases from 0.610 and 0.622 to 0.659 and 0.660 for low and high resistance, while separation increases from 0.219 and 0.245 to 0.318 and 0.320. Group-level alignment increases from 0.669 and 0.688 to 0.716 and 0.716, while separation increases from 0.339 and 0.376 to 0.432 and 0.432. These shifts are large relative to the bounded scale of MAS. They confirm that training sharpens class-consistent evidence at the component level, which directly strengthens the composed property prototypes and enlarges retrieval margins.

\begin{table}[htbp]
\centering
\caption{MAS summary for the AgNO$_3$ 10\% holdout. Mean alignment and separation are averaged over components within each class partition. Values are reported at initialization, at convergence, and as the change from initialization to convergence.}
\label{tab:mas_summary_results}
\small
\setlength{\tabcolsep}{3.5pt}
\begin{tabular}{lccccccc}
\toprule
 & \multirow{2}{*}{Metric} &
\multicolumn{3}{c}{Low resistance} &
\multicolumn{3}{c}{High resistance} \\
\cmidrule(lr){3-5}\cmidrule(lr){6-8}
 &  & Epoch 0 & Final & Change & Epoch 0 & Final & Change \\
\midrule
Param. & $\mathrm{MA}_c$ & 0.610 & 0.659 & +0.049 & 0.622 & 0.660 & +0.038 \\
-level & $\mathrm{MS}_c$ & 0.219 & 0.318 & +0.099 & 0.245 & 0.320 & +0.075 \\
\midrule
Group & $\mathrm{MA}_c$ & 0.669 & 0.716 & +0.047 & 0.688 & 0.716 & +0.028 \\
-level & $\mathrm{MS}_c$ & 0.339 & 0.432 & +0.093 & 0.376 & 0.432 & +0.056 \\
\bottomrule
\end{tabular}
\end{table}

\subsubsection{Sample-level behavior} \label{sec: Sample level}
\begin{figure}[htbp]
  \centering
  \includegraphics[width=0.8\linewidth]{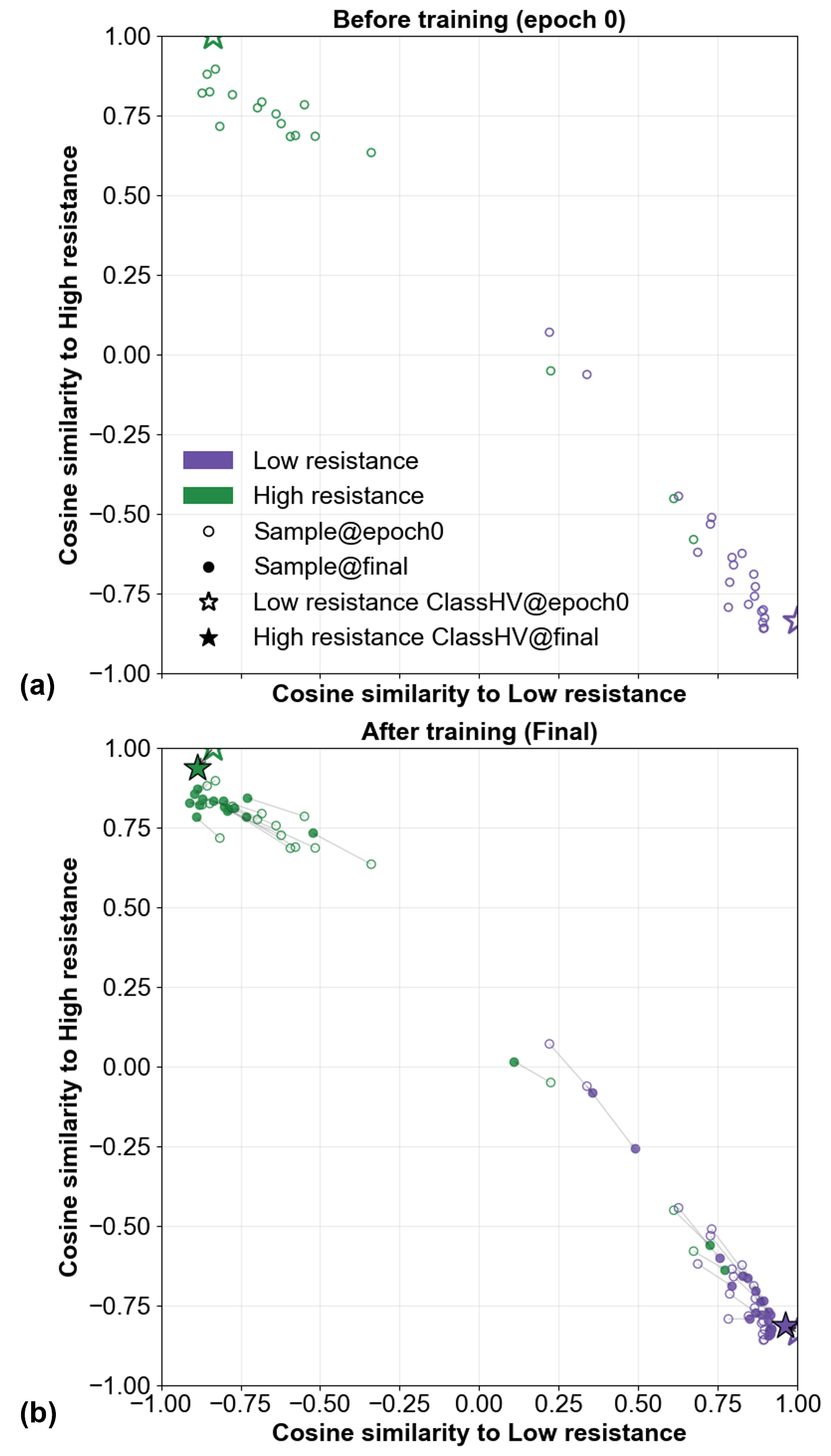}
  \caption{Training-set sample hypervectors embedded by cosine similarity to the two epoch-0 class prototypes. The panel (a) shows initialization and the panel (b) shows the post-training configuration mapped back to the same coordinate system. Light gray segments indicate sample displacement. Star markers denote class prototypes. Samples concentrate toward their correct prototype and inter-class overlap decreases after training.}
  \label{fig:sample_move}
\end{figure}

We next visualize how training reshapes sample hypervectors. Each training sample is embedded into a two-dimensional coordinate system given by its cosine similarities to the two epoch-0 class prototypes. For a sample hypervector $\mathbf{h}$, we plot \(
\big(\operatorname{sim}(\mathbf{h},\mathbf{m}_1^{(0)}),\ \operatorname{sim}(\mathbf{h},\mathbf{m}_2^{(0)})\big),
\), where $\mathbf{m}_1^{(0)}$ and $\mathbf{m}_2^{(0)}$ denote the two class prototypes at initialization. Figure~\ref{fig:sample_move} shows the sample distribution at epoch 0 and after convergence. In the final panel, similarities are computed to the final prototypes and mapped back into the epoch-0 coordinate system for direct comparison. Light gray segments indicate each sample displacement.
Most samples move toward their correct prototype and the class clusters become more compact and better separated. A small number of outliers remain far from their true prototype, consistent with non-perfect training accuracy in Fig.~\ref{fig:dyn_global}(b). This sample movement provides direct geometric evidence that the adaptive encoder improves class separation by reshaping sample representations rather than relying only on prototype drift. We quantify this effect using cosine distance to the correct prototype $d(\mathbf{h},\mathbf{m}_{y})=1-\operatorname{sim}(\mathbf{h},\mathbf{m}_{y})$, which is proportional to squared Euclidean distance for unit-normalized hypervectors since $\|\mathbf{h}-\mathbf{m}_{y}\|_2^2=2d$. 

Table~\ref{tab:proto_dist_detailed} reports mean and standard deviation of $d$ on both train and test sets at epoch 0 and after convergence. The mean distance decreases on the training set from $0.320\pm0.336$ to $0.229\pm0.390$, and it also decreases on the test set from $0.645\pm0.326$ to $0.537\pm0.401$. The concurrent reduction on both splits confirms that the learned encoder increases prototype alignment in a manner that generalizes, rather than simply memorizing the training samples.

\begin{table*}[htbp]
\centering
\caption{Mean and standard deviation of cosine distance to the correct prototype, $d(\mathbf{h},\mathbf{m}_y)=1-\operatorname{sim}(\mathbf{h},\mathbf{m}_y)$, reported for train and test sets at epoch 0 and after convergence. Change is final minus epoch 0, where negative indicates movement toward the correct prototype.}
\label{tab:proto_dist_detailed}
\scriptsize
\setlength{\tabcolsep}{3.5pt}
\resizebox{\linewidth}{!}{%
\begin{tabular}{lccc ccc ccc}
\toprule
& \multicolumn{3}{c}{Overall} & \multicolumn{3}{c}{Low resistance} & \multicolumn{3}{c}{High resistance} \\
\cmidrule(lr){2-4}\cmidrule(lr){5-7}\cmidrule(lr){8-10}
Split & Epoch 0 & Final & Change
      & Epoch 0 & Final & Change
      & Epoch 0 & Final & Change \\
\midrule
Train
& 0.320$\pm$0.336 & 0.229$\pm$0.390 & $-$0.091$\pm$0.076
& 0.233$\pm$0.176 & 0.131$\pm$0.153 & $-$0.101
& 0.422$\pm$0.435 & 0.342$\pm$0.528 & $-$0.080 \\
Test
& 0.645$\pm$0.326 & 0.537$\pm$0.401 & $-$0.107$\pm$0.126
& 0.532$\pm$0.277 & 0.452$\pm$0.367 & $-$0.081
& 0.888$\pm$0.291 & 0.723$\pm$0.411 & $-$0.165 \\
\bottomrule
\end{tabular}%
}
\end{table*}

\subsection{PSP-HDC Explainability}\label{sec: PSP-HDC Explainability}
Having established that PSP-HDC is the most accurate and most consistent method under both evaluation protocols, we now expose its decisions through prototype retrieval on the PSP graph. Unless otherwise stated, attribution results are computed under the 1000 random splits and then averaged across splits. For each split, we compute parameter-level, group-level, and within-group attributions from class-partitioned memories. Averaging across splits suppresses sampling noise and yields stable, split-robust explanations.

\subsubsection{Parameter-level attribution}
\begin{figure}[htbp]
  \centering
  \includegraphics[width=\linewidth]{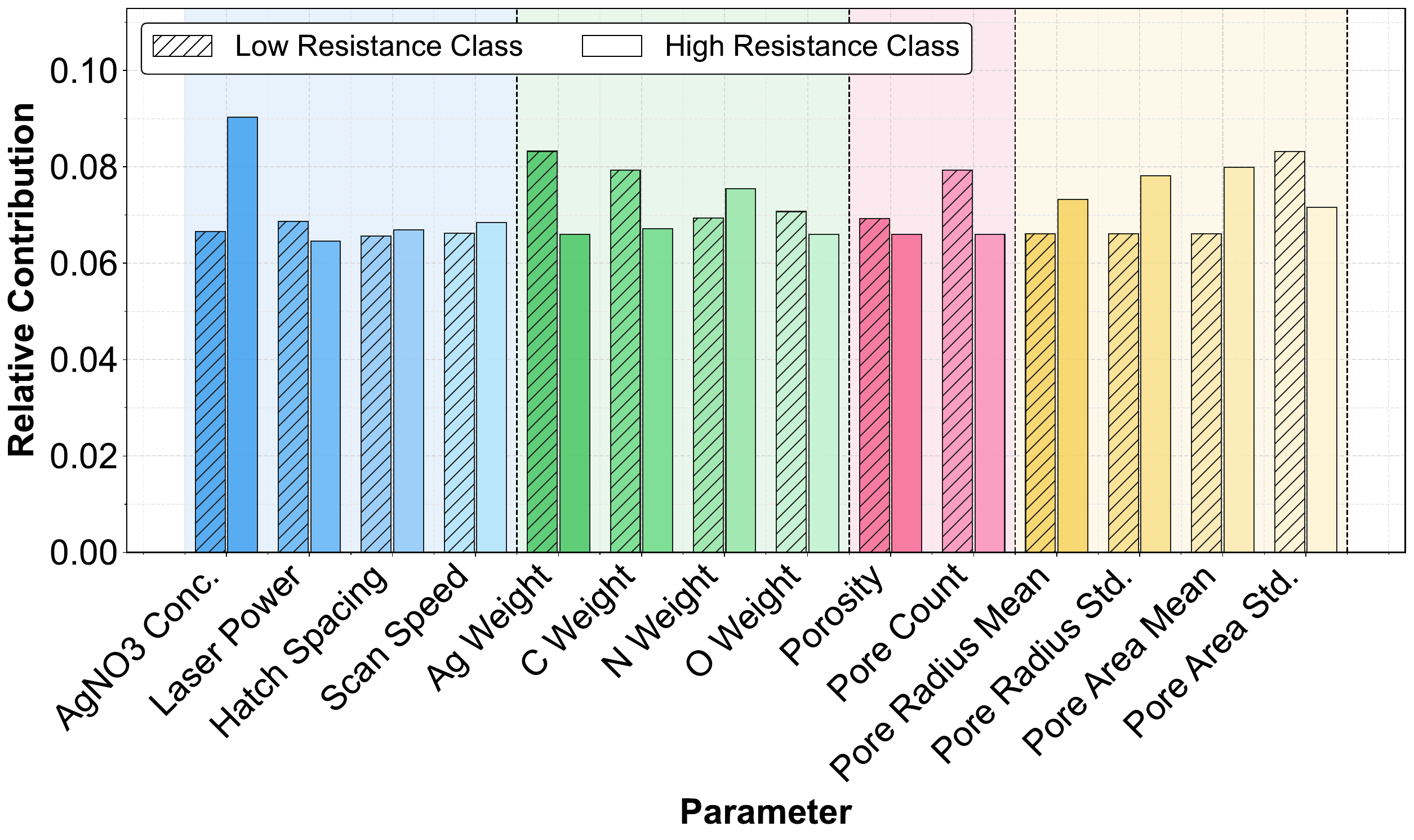}
  \caption{Parameter-level attribution for PSP-HDC. Bars show the class-partitioned relative contribution of each scalar parameter to the prototype-based decision for low and high resistance. The dashed vertical line separates process parameters from structure parameters including elemental composition and morphological descriptors.}
  \label{fig:Parameter-level}
\end{figure}

Figure~\ref{fig:Parameter-level} reports class-partitioned parameter attributions. No single parameter dominates across both classes, consistent with the coupled PSP mapping shown in Section~\ref{sec: Preliminary Analysis}. The two classes nonetheless emphasize different evidence. High-resistance decisions allocate larger weight to AgNO$_3$ concentration among process variables together with pore-size dispersion descriptors on the structure side. Low-resistance decisions distribute weight more strongly toward composition-related descriptors together with morphology, indicating that conductive predictions are supported by joint chemical and microstructural evidence rather than by a single control variable.

\subsubsection{Parameter-group attribution}
\begin{figure}[htbp]
  \centering
  \includegraphics[width=\linewidth]{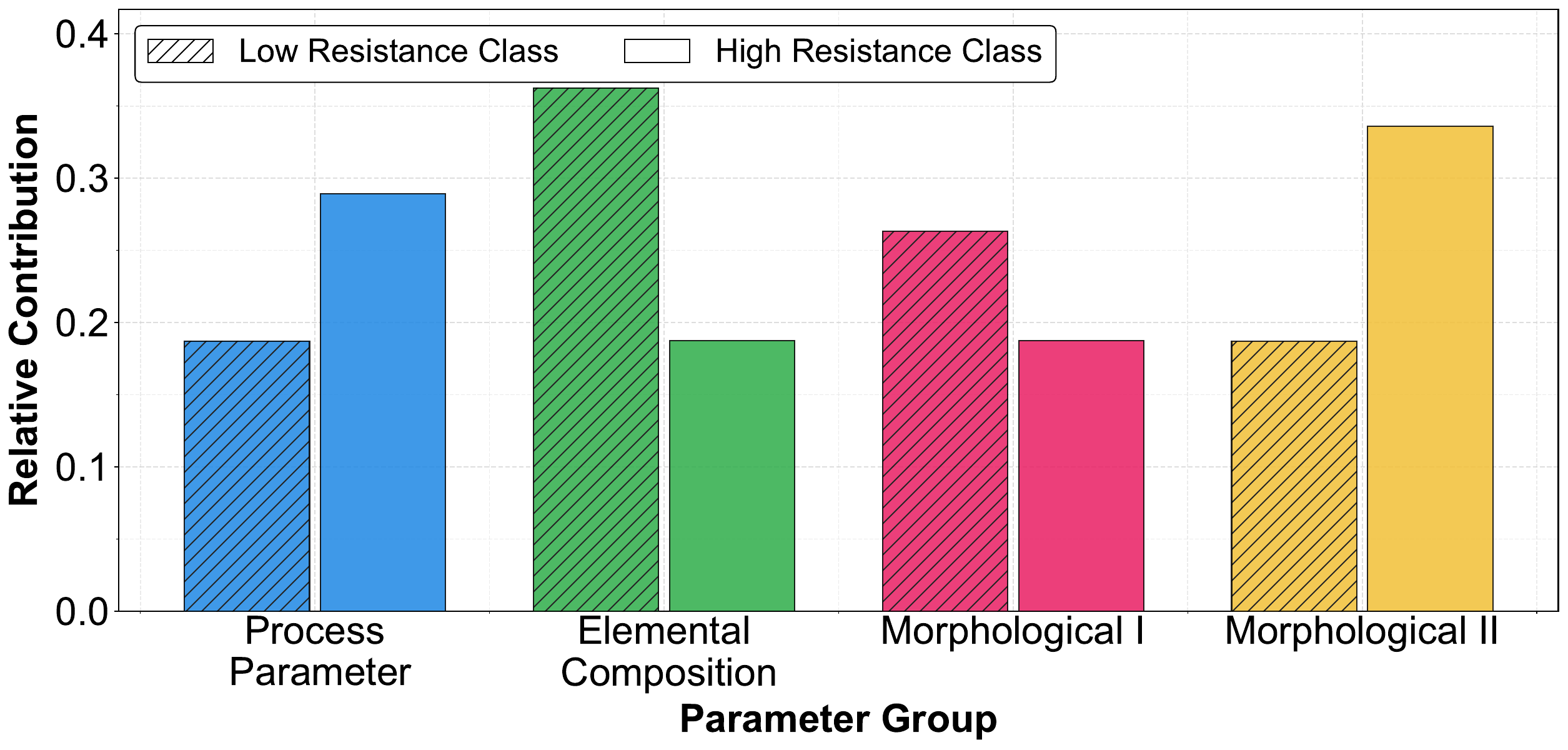}
  \caption{Parameter-group attribution on the PSP graph. Bars show the relative contribution of each parameter group to the prototype-based decision for low and high resistance, revealing which PSP regions dominate the evidence through combined effects.}
  \label{fig:Group-level}
\end{figure}

Group-level attributions expose which regions of the PSP graph dominate the class evidence. Figure~\ref{fig:Group-level} shows a clear class-conditional re-weighting. For low resistance, the dominant group is elemental composition followed by the first morphology group, indicating that the model justifies conductivity primarily through composition and coarse pore configuration. For high resistance, the dominant group shifts to the second morphology group and the process group increases its share, indicating that insulating outcomes are justified by pore-geometry dispersion signatures together with process conditions. This is precisely the type of dependency-aware explanation that feature-centric post hoc rankings cannot provide under correlated PSP variables.

\subsubsection{Within-group attribution}
\begin{figure}[htbp]
  \centering
  \includegraphics[width=\linewidth]{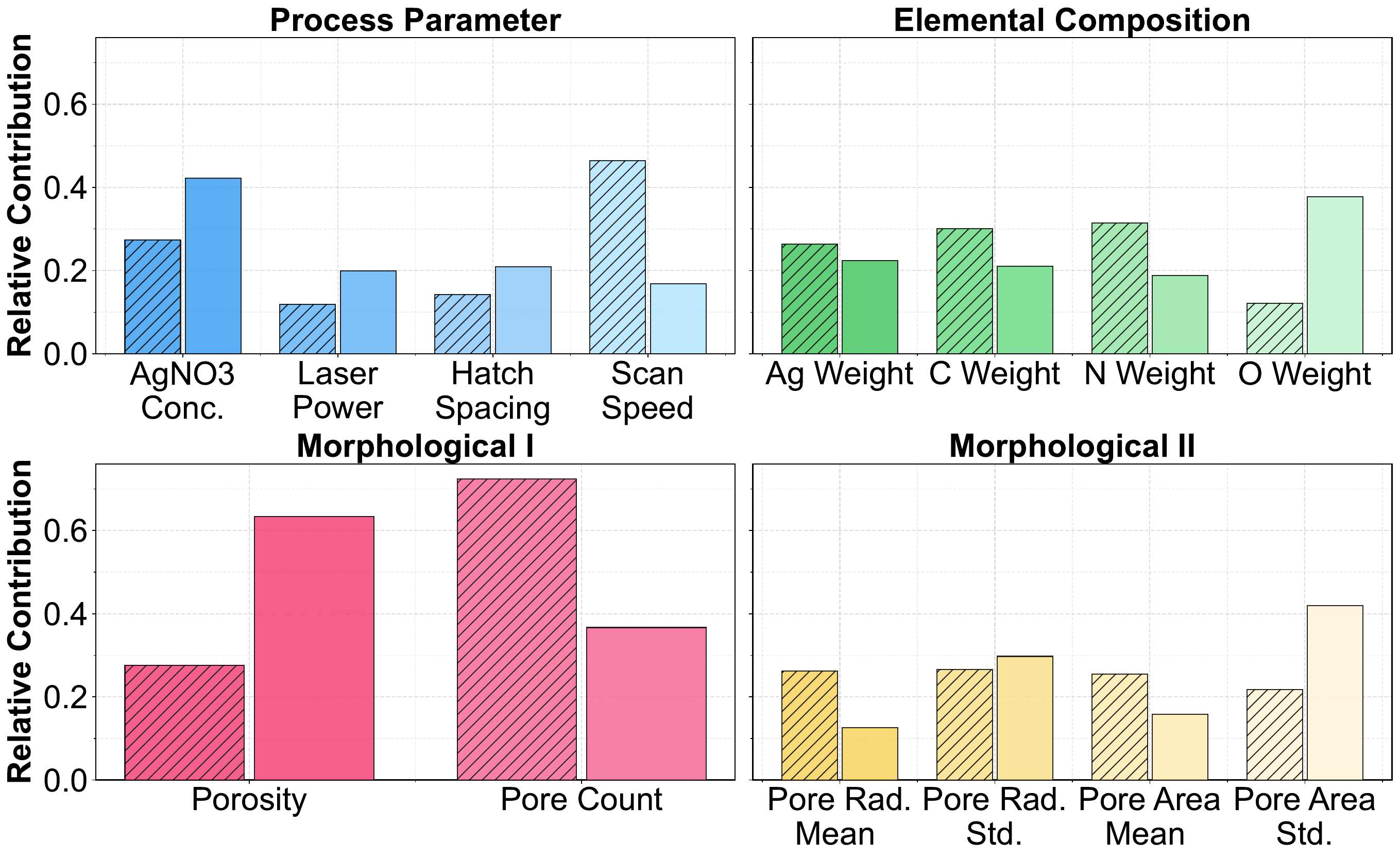}
  \caption{Within-group attribution decomposition. Each panel corresponds to one parameter group. Bars show the relative contributions of individual parameters within that group, normalized within the group, for low and high resistance classes.}
  \label{fig:Within-group}
\end{figure}

Within-group decomposition identifies the drivers inside each group. Figure~\ref{fig:Within-group} reports attributions normalized within each group, so they quantify relative contribution inside the group rather than global importance.

Low-resistance evidence is driven primarily by scan speed, consistent with the strong speed dependence in Fig.~\ref{fig:rs_overview}(b). High-resistance evidence places greater weight on AgNO$_3$ concentration, consistent with the prevalence of insulating outcomes at low concentration and the concentration-dependent shifts in Fig.~\ref{fig:rs_overview}(a).

High-resistance evidence is dominated by oxygen weight, while low-resistance evidence is distributed more broadly across Ag, C, and N with reduced emphasis on oxygen. This indicates that oxygen-related composition signatures are a key discriminator supporting high-resistance decisions.

Within the porosity and pore-count group, low-resistance evidence emphasizes pore count while high-resistance evidence emphasizes porosity. This validates the need for grouped modeling because similar porosity values can arise from distinct pore configurations and carry different implications.

High-resistance evidence is driven by dispersion statistics such as pore area standard deviation and pore radius standard deviation, while low-resistance evidence is more balanced between mean and dispersion descriptors. This indicates that high resistance is justified by heterogeneity and irregularity in pore geometry rather than by average pore size alone.

\subsubsection{Why attributions differ across classes}
Class-conditional attribution differences are intrinsic to PSP-HDC. The model stores class-partitioned component memories and predicts by retrieval against class prototypes. Attribution therefore quantifies which components increase the similarity margin of one class prototype relative to its competitor, rather than producing a single class-agnostic ranking. In addition, the two classes occupy different regions of the PSP table with different dispersion and correlation structure across variables. Evidence that is discriminative for low resistance is not required to mirror the evidence that is discriminative for high resistance. Differences across attribution levels are also structural. Parameter-level attribution measures global per-parameter contribution to class evidence. Group-level attribution measures joint effects induced by bundling within PSP graph nodes. Within-group attribution resolves how individual parameters compose the group evidence. Together, these levels provide a complete, graph-aligned explanation of the decision logic under coupled PSP interactions.

\section{Conclusion} \label{sec: Conclusion}

This paper presented PSP-HDC, a graph-based hyperdimensional computing framework for per-sample process--structure--property (PSP) prediction in a data-scarce multiphoton photoreduction setting. PSP-HDC operationalizes a domain-defined directed PSP graph as an internal structural prior: heterogeneous scalar process and structure descriptors are encoded as hypervectors and composed along PSP-graph structure via graph-aligned binding and bundling, rather than being treated as an unstructured feature vector. Property prediction is performed through similarity-based associative-memory retrieval against learned class prototypes, yielding an explicit, memory-centric decision mechanism with a measurable similarity margin.
To improve learning on small, heterogeneous PSP tables, PSP-HDC incorporates a gradient-based adaptive scalar-to-hypervector encoder that learns parameter-specific embeddings on a fixed high-dimensional basis, increasing representational flexibility while preserving the efficiency and robustness of HDC. Beyond prediction, PSP-HDC provides intrinsic, PSP-graph-aligned explainability by decomposing class evidence at three complementary levels: parameter-level contributions, parameter-group effects induced by PSP-graph bundling, and within-group decompositions that resolve how individual parameters shape group evidence. We further introduced memory alignment and separation and complementary learning-dynamics analyses to characterize how training reshapes prototype geometry and sharpens class separation.
We evaluated PSP-HDC on the 3D OHMIC platform while restricting the study to specimens fabricated using its multiphoton photoreduction capability. PSP-HDC achieved $0.910 \pm 0.077$ accuracy over 1000 random splits and $0.896$ under process-fold generalization, outperforming the evaluated baselines on overall performance while maintaining transparent, dependency-aware explanations aligned with directed PSP structure. These results show that graph-structured HDC can recover transferable PSP coupling from small, heterogeneous experimental datasets rather than merely fitting regime-specific correlations.
Future work will extend PSP-HDC to richer PSP objectives, including multi-class and continuous property prediction, explicit treatment of right-censored measurements, and uncertainty-aware decision rules for reliable deployment. We will also investigate adaptive PSP-graph refinement and experiment-selection strategies to accelerate data-efficient process development in photoreduction-driven manufacturing workflows.

\section*{Acknowledgments}
This work was supported under Cooperative Agreement W56HZV-21-2-0001 with the US Army DEVCOM Ground Vehicle Systems Center (GVSC), through the Virtual Prototyping of Autonomy Enabled Ground Systems (VIPR-GS) program, and by the National Science Foundation, United States (Grant No. 2434519).

DISTRIBUTION STATEMENT A. Approved for public release; distribution is unlimited. OPSEC10422

Disclaimer: Reference herein to any specific commercial company, product, process, or service by trade name, trademark, manufacturer, or otherwise, does not necessarily constitute or imply its endorsement, recommendation, or favoring by the United States Government or the Department of the Army (DoA). The opinions of the authors expressed herein do not necessarily state or reflect those of the United States Government or the DoA and shall not be used for advertising or product endorsement purposes.

\bibliographystyle{IEEEtran}
\bibliography{cas-refs}


 


\vfill

\end{document}